\documentclass[sigconf]{acmart}
\usepackage{graphicx}
\usepackage{subfigure,amsmath}
\usepackage{booktabs} 
\usepackage{tikz}
\usetikzlibrary{fit,arrows,backgrounds,shapes}
\usepackage{hyperref,url,defs}
\RequirePackage[textsize=scriptsize]{todonotes}

\begin{document}

\copyrightyear{2019} 
\acmYear{2019} 
\setcopyright{acmcopyright}
\acmConference[KDD '19]{The 25th ACM SIGKDD Conference on Knowledge Discovery and Data Mining}{August 4--8, 2019}{Anchorage, AK, USA}
\acmBooktitle{The 25th ACM SIGKDD Conference on Knowledge Discovery and Data Mining (KDD '19), August 4--8, 2019, Anchorage, AK, USA}
\acmPrice{15.00}
\acmDOI{10.1145/3292500.3330996}
\acmISBN{978-1-4503-6201-6/19/08}

\settopmatter{printacmref=true}
\fancyhead{}


\title{Streaming Adaptation of Deep Forecasting Models using Adaptive Recurrent Units}



\author{Prathamesh Deshpande}
\affiliation{IIT Bombay}
\email{pratham@cse.iitb.ac.in}

\author{Sunita Sarawagi}
\affiliation{IIT Bombay}
\email{sunita@iitb.ac.in}

\newcommand{\vg}{{\vek{g}}}
\newcommand{\ModifiedARU}{ARU}


\begin{abstract}
We present ARU, an Adaptive Recurrent Unit for streaming adaptation of
deep globally trained time-series forecasting models.  The ARU
combines the advantages of learning complex data transformations across
multiple time series from deep global models, with per-series localization offered by 
closed-form 
linear models.  Unlike existing methods of adaptation that are either
memory-intensive or 
non-responsive 
after training, ARUs require only fixed sized state and adapt 
to streaming data via an easy 
RNN-like 
update operation.  
The core principle driving ARU is simple --- maintain sufficient statistics of conditional Gaussian distributions and use them to compute local parameters in closed form.  Our contribution is in embedding such local linear models in globally trained deep models while allowing end-to-end training on the one hand, and easy RNN-like updates on the other.  Across several datasets we show that ARU is more effective  than recently proposed local adaptation methods that tax the global network to compute local parameters.
\end{abstract}



\maketitle

\section{Introduction}
Time series forecasting is a critical analysis tool in retail~\cite{larson01,johnson15}, finance, and commerce, and has been extensively researched across communities for multiple decades~\cite{box64,box68,hyndman08,quinlan92,Seeger2016,FlunkertSG17,Araujo:2018} (See \cite{Faloutsos2018} for a recent survey).
However, with the remarkable success of deep learning in other classically hard tasks like speech recognition, the time series forecasting problem also requires revisiting.  
We consider typical forecasting settings where we are given multiple series along different points in time. Each series at time $t$ is characterized by a real-valued output $y^i_t$ and a vector of input features $\vx^i_t$.  Our goal is to predict the $y$  for $K$ future points of time for which we are given inputs $\vx^i_{T+1}, \ldots, \vx^i_{T+K}$.

Classical methods of time series analysis like ARIMA, exponential smoothing, and other state space models~\cite{hyndman08}, train separate parameters $\vtheta^i$ for each time series $i$ and can be considered as local models.  
In contrast recent deep learning methods~\cite{FlunkertSG17,wen2017multi,Mukherjee2018} train shared parameters $\vtheta$ for predicting each $y$ as a function of a summary of each time series' previous observation.  
Typically, a Recurrent Neural Networks (RNNs) is used  for representing such sequential history as fixed dimensional vectors. For sequences representing time series, in addition to general purpose gated RNNs like LSTMs, recent proposals also include un-gated RNNs like Statistical Recurrent Units (SRUs~\cite{OlivaPS17}) and Fourier Recurrent Units (FRU~\cite{ZhangLSD18}).  During training, data from multiple time series teaches the RNN to capture enough context from the history relevant for making predictions at 
each of the future $K$ points in time.
Such models have been found to surpass purely local models like ARIMA and conventional globally trained models like boosted regression trees~\cite{quinlan92}. 

In general however such global model may be sub-optimal when the various series are heterogeneous, or when not all time series are observed during training.  The only local influence in these models is the RNN state 
that evolves with time separately for each time series $i$. 
%
%
We take a different view and treat the per-series state as  {\em local parameters} that map a future input along with its context to a prediction. We use a clever trick to update the local parameters in closed form, and in a streaming manner much like normal RNN parameters.  We call our unit a Adaptive Recurrent Unit (ARU) whose finite-sized state 
can be used to compute 
per-series local parameters in closed form.

Recently, other methods have also been proposed to augment a global time series model with local parameters\cite{Rangapuram2018,Goel2017}.
However, their method of local adaptation requires the global model to be retrained for each update to a time series. 
More generally, localization of a globally trained model can be treated as a domain adaptation problem~\cite{bendavid06Analysis,BlitzerMP06} for which many methods have been proposed including model fine-tuning, meta-learning~\cite{Finn2017,ravi17,mishra2018a}, and memory-augmented networks~\cite{SantoroBBWL16,Shankar2018}.  These methods 
either require gradient-based iterative updates, or are memory-intensive, and/or rely on techniques like self-attention that is 
quadratic in $T$ for large time-series.
ARU does not entail iterative training of local parameters, its adaptation is light-weight and streaming, and only requires a time-series specific state of constant size.
This makes it particularly efficient for long time-series since its storage and compute requirement is constant in the sequence length. 

The core principle of ARU is to use RNN-like updates to incrementally maintain per-series sufficient statistics required for fitting a local conditional Gaussian distribution in closed form.  Our contribution is in exploiting this classical trick in the context of modern deep learning models, and designing network architectures that provide the right local-global tradeoff.  Surprisingly, we show that this strategy of local adaptation is more robust across different types of time-series than recently proposed more ambitious methods that tax the deep network to learn to generate local parameters from RNN states.

We compare our ARU-based adaptation to two state-of-art adaptation methods on five time-series datasets under various settings.  We show that ARU is effective in reducing error by 10 to 20\% compared to the baseline, and is generally better than existing adaptation methods.   This reduction in error comes at very little overheads of running time, whereas existing state-of-the-art adaptation methods are up to a factor of four slower.  








\newcommand{\cN}{{\mathcal N}}
\renewcommand{\vv}{{\vx}}
\newcommand{\vl}{{\vek{s}}}
\section{Our Model}

We start with a review of state-of-the-art global models and classical local models for time series forecasting.
\subsubsection*{Review: Global Models}
Typical deep learning based global models for multi-horizon time series forecasting~\cite{FlunkertSG17,wen2017multi} deploy the encoder-decoder architecture.   
First an input layer maps the input features $\vx^i_t$ to a real vector. 
This could include embedding lookups for categorical attributes and any rescaling for continuous attributes. Next the transformed input along with the previous 
output 
$[\vv^i_t, y_{t-1}^i]$ is  fed to one or more encoder RNN layers.  
The RNN could be either a generic LSTM or time-series specific units like SRU or FRU that capture 
context useful for future predictions. 
The output of the encoder is its final state $\vg^i_T$ at the end of $T$ steps. This can be treated as a summary of the known $y$ values that is relevant as a {\em context} for future predictions.    
The decoder initialized with $\vg^i_T$ is responsible for making the predictions on the $K$ future time horizons as a function of respective inputs $\vx^i_{t}: t=T+1\ldots,T+K$.  The decoder could be auto-regressive, 
where 
the previous predicted $y$ is fed as input to be next step, or independently make each of the $K$ predictions. We found the independent model to provide higher accuracy for long-term forecasts than the auto-regressive model fed with noisy previous predictions. This is corroborated in \cite{wen2017multi}.
The decoder
takes each transformed future input $\vv^i_{t}: t=T+1\ldots,T+K$ concatenated with $\vg^i_T$ and generates an output vector $\vh^i_t$  using an optional RNN and one or more feed-forward layers.  Lastly, a Gaussian distribution is imposed on the output by using a linear layer to map $\vh^i_t$ into a mean and variance. The final equations driving the prediction at future times are: 
\begin{eqnarray}
  \label{eq:globalmu}
& &\vg^i_T  = RNN([y^i_{t-1},\vx^i_t]:t=1\ldots T|\theta_{\text{enc}}) \\
& & \vh^i_t  = FF([\vg^i_T, \vx^i_t]:t=T+1\ldots T+K | \theta_{\text{dec}}) \\ 
& &\mu^i_t  =\theta_\mu[\vh^i_t,1],~~~\sigma^i_t = \log(1+\exp(\theta_\sigma[\vh^i_t,1])) \\
& &\Pr(y^i_t | \vx^i_t, (\vx^i_1,y^i_1),\ldots, (\vx^i_T,y^i_T)) = \cN(\mu^i_t, \sigma^i_t)
\end{eqnarray}
where $\theta_{\text{enc}}, \theta_{\text{dec}}, \theta_\mu, \theta_\sigma$ are all parts of the global parameters $\vtheta$.
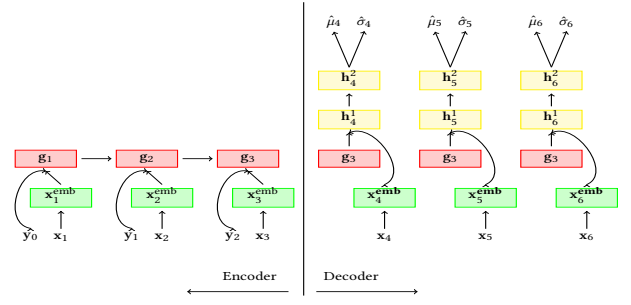
\begin{figure}
    \centering
    \resizebox{8.3cm}{3.91cm}{
\begin{tikzpicture}[
  hid/.style 2 args={
    rectangle split,
    rectangle split horizontal,
    draw=#2,
    rectangle split parts=#1,
    fill=#2!20,
    outer sep=1mm,
    text width=1cm, 
    text height=1.5mm, 
    text centered}]
  \foreach \t [count=\step from 4] in {{{$\hat{\mu}_4$}},{{$\hat{\mu}_5$}},{{$\hat{\mu}_6$}}} {
    \node[align=center] (om\step) at (2*\step-0.3, 3.50) {\t};
  }
  \foreach \t [count=\step from 4] in {{{$\hat{\sigma}_4$}},{{$\hat{\sigma}_5$}},{{$\hat{\sigma}_6$}}} {
    \node[align=center] (os\step) at (2*\step+0.3, 3.50) {\t};
  }
  \foreach \step in {1,...,3} {
    \pgfmathtruncatemacro\prev{\step + -1}
    \node[hid={1}{red}] (h\step) at (2*\step, 0) {{{$\vg_\step$}}};
    \node[hid={1}{green}] (e\step) at (2*\step+0.3, -1) {{{$\vv_\step^{\text{emb}}$}}};
    \node[] (yprev\step) at (2*\step-0.3, -2) {{{$\vy_\prev$}}};
    \node[] (inp\step) at (2*\step+0.3, -2) {{{$\vv_\step$}}};
    \draw[->] (yprev\step.north) edge[bend left = 90] (h\step.south);
    \draw[->] (inp\step.north) -> (e\step.south);
    \draw[->] (e\step.north) -> (h\step.south);
  }
  \foreach \step in {4,...,6} {
    \node[hid={1}{yellow}] (h2\step) at (2*\step, 2) {{{$\vek{h}^2_\step$}}};
    \node[hid={1}{yellow}] (h1\step) at (2*\step, 1) {{{$\vek{h}^1_\step$}}};
    \node[hid={1}{red}] (h\step) at (2*\step, 0) {{{$\vg_3$}}};
    \node[hid={1}{green}] (e\step) at (2*\step+0.7, -1) {{{$\vv_\step^{\textbf{emb}}$}}};
    \node[] (inp\step) at (2*\step+0.7, -2) {{{$\vv_\step$}}};
    \draw[->] (inp\step.north) -> (e\step.south);
    \draw[->] (e\step.north) edge[bend right = 90] (h1\step.south);
    \draw[->] (h\step.north) -> (h1\step.south);
    \draw[->] (h1\step.north) -> (h2\step.south); 
    \draw[->] (h2\step.north) -> (om\step.south);
    \draw[->] (h2\step.north) -> (os\step.south);
  }  
  \foreach \step in {1,...,2} {
    \pgfmathtruncatemacro{\next}{add(\step,1)}
    \draw[->] (h\step.east) -> (h\next.west);
  }
  \foreach \step in {4,...,6} {
    \pgfmathtruncatemacro{\next}{add(\step,1)}
  }
  
  \draw (7.1,-3.5) -- (7.1,4);
  \node[text width=4cm] at (7.5,-3) {Encoder};
  \node[text width=4cm] at (9.5,-3) {Decoder};
  \draw[->] (6.8,-3.4) -- (4.8,-3.4);
  \draw[->] (7.35,-3.4) -- (9.35,-3.4);
\end{tikzpicture}
}
    \caption{Diagram of the global model with encoder size of $3$ and decoder size of $3$. All the nodes/states with the same label are copies.}
    \label{Fig:global_model}
\end{figure}
During training we are given data comprising of 
input-output
pairs over several time series. $D=\{(\vx^i_1,y^i_1),\ldots,(\vx^i_T,y^i_T): i = 1 \ldots, N\}$. We simulate multiple encoder-decoder windows from this history using a sliding window of stride $\ell$ and compute the data likelihood as follows:
$$\max_\vtheta \sum_{i=1}^N \sum_{T'=2:\ell}^{T-K}\sum_{t=T'+1}^{T'+K}\log {\mathcal N}(y^i_t ; \mu^i_t, \sigma^i_t | \vtheta)$$
where ${\mathcal N}$ denotes the Gaussian density function. The parameter $\vtheta$ includes all network parameters spanning the input-layer, the encoder RNN, the decoder RNN, and the last output layer that generates  $\mu^i_t, \sigma^i_t$. Figure \ref{Fig:global_model} shows the global model.

The global model is driven by parameters trained across multiple
time-series, and the only local influence in the final equation that
outputs 
$\Pr(y^i_t|\vx^i_t, \lbrace(\vx^i_t,y^i_t)\rbrace_{t=1}^{T})$
is the RNN state $\vg^i_T$ that serves as
context. The RNN parameters are trained end-to-end to find the most
relevant context.

\subsubsection*{Review: Local Models}
A local model would train separate parameters $\theta^i$ for each $i$.
However, unless the length of each time-series is very large, training
parameters of a complex network like a multi-layer deep network would
not work.  Hence, local models have traditionally been simple models
such as linear state space models that use linear parameters to
transition from one state to the next, and make state-ful predictions.  Let $\vs_t^i$ denote the local state at time $t$ of series $i$.  Parameters $\theta^i_{\text{tr}}$ control transition from one state to the other via affine transforms, and parameter $\theta^i_\mu$ controls the linear transform to generate the mean output from the local state. The variance is a fixed learned parameter, and that makes the local model as:
\begin{eqnarray}
  \label{eq:local}
  \vl^i_t & = \theta^i_{\text{tr}} [\vl^i_{t-1}~ \vx^i_t ~1] \\
  \mu^i_t & =\theta^i_\mu[\vl^i_t~1],~~~\sigma^i_t=\theta^i_\sigma
\end{eqnarray}
where the parameters $\theta^i_{\text{tr}}$ and $\theta^i_\mu,
\theta^i_\sigma$ are learned locally for each time-series using its known values up to $T$ that is $(\vx^i_1,y^i_1),\ldots, (\vx^i_T,y^i_T)$.  Assuming $T$ is large, local
models can capture peculiarities of individual time-series in ways
that cannot be approximated by local RNN state of global models.


\subsection{The ARU RNN}
Our model not only attempts to capture the best of the local and global models,
but does so in a manner that allows the local parameters to be
computed recurrently in a streaming manner.  We call our unit Adaptive
Recurrent Unit (ARU) that views the per-series state as local
predictive parameters that are fitted on the fly.
Additionally, ARU
is designed to capture local predictive parameters without expensive
parameter retraining.

We use the shared global network to learn the complex transformation of the history and the input into a final output vector $\vh^i_t$. However, instead of the global last linear layer (Equation~3 in the global model), we use a local model.  We perform the local fit in a clever way to allow easy end-to-end training of the global network parameters without inner loop of local updates as in existing methods like \cite{Finn2017}.  
The main idea behind the ARU is to use a RNN to provide the best possible least square fit locally for each time series based on observed $\vx_t,\vh_t$ values.  
The motivation behind this choice is that the optimal local parameters can be obtained in closed form using sufficient statistics that can be maintained in a streaming manner.  In contrast local parameters that depend on iterative gradient-based updates, require multiple data passes and are difficult to embed within a larger global network that is trained via gradient descent.
Although the ARU unit itself only models linear interaction between
its inputs and outputs, it is embedded in a larger neural network that
provides non-linear transformation of both the input to the ARU and
its output. 
We describe the ARU-RNN in this section
and then in Section~\ref{sec:aru_in_global_model} describe how we fit
it in the global model.


\newcommand{\sxx}{\vek{sxx}^{i}}
\newcommand{\sxy}{\vek{sxy}^{i}}
\newcommand{\ssigma}{\vek{ss}^{i}}
\newcommand{\sn}{\vek{sn}^{i}}

\subsubsection{The ARU Update Equations}
Let $\vh^i_t$ be the output vector at time $t$ from the global model.
The ARU RNN has two modes: adapt and predict. In the adapt mode it is fed the input $\vh^i_t$ along with the true label $y^i_t$ at that time and produces an updated state $\vs^i_t$.  In the predict mode denoted $ARU(\vs^i_{t-1}, \vh_t^i)$ it outputs the local prediction based on state  $\vs^i_{t-1}$ on the input  $\vh_t^i$. 

The ARU state $\vs_t$ at each time $t$ keeps four types of sufficient statistics $\vs_t=[\sxx_t,\sxy_t,\sn_t,\ssigma_t]$ to provide a local least square fit between the $\vh$ and $y$s. A vector $\valpha$ of $J$ aging factors maintains this statistics for varying amounts of aging of the old data. Initially,  all states are zero and at each time-step it is updated as follows:

ARU\_Update($\vek{s}^i_{t-1}, \vh^i_t, y^i_t):$
\begin{align*}
\sxx_t &= \valpha\,\sxx_{t-1} + [\vh_t^i~1]^{\text{T}}[\vh_t^i~1] \\
\sxy_t &= \valpha\,\sxy_{t-1} + [\vh_t^i~1]^{\text{T}}(\vy_t^i) \\
\sn_t &= \valpha\,\sn_{t-1} + 1 \\
\ssigma_t &= \valpha\,\ssigma_{t-1} + (\vy_t^i-ARU(\vs^i_{t-1}, \vh_t^i))^2 \\
\text{Return} &~\vs^i_t = [\sxx_t,\sxy_t,\sn_t,\ssigma_t]
\end{align*}
From these update equations, it is easy to interpret what each of the ARU state components represent.  
The state $\sxx$ represents age-weighted sum of pairwise  feature product of the input vector, $\sxy$ represents age-weighted sum of product of input features and output $y$,  $\sn_t$ represents the age adjusted count so far, and  $\ssigma_t$ represents the accumulated fitting noise that will serve as the variance from the local predictions.  In this equation $ARU(\vs^i_{t-1}, \vh_t^i)$ denotes the mean local prediction as explained below.
 The ARU state is updated in a streaming manner every time a true $y^i_t$ is known.

\sloppy We next provide the equations of ARU in the predict mode. $ARU(\vs^i_{t-1}, \vh_t^i)$ represents the predicted values from the ARU at state $\vs^i_{t-1}$ on the input  $\vh_t^i$.
The predicted output using only ARU states can be computed by exploiting the  closed form of the least squares solution. First, the local parameter is calculated in closed form using the sufficient statistics. The $\lambda I$ term regularizes the local parameters. The local parameter is used to compute the mean and variance of the local prediction.

ARU\_Predict($\vek{s}^i_t, \vh^i_t$):
\begin{alignat}{2}
\label{Eqn: ARU_Predict}
\vtheta^i_{t,\mu} & = (\sxx_t + \lambda I)^{-1}\sxy_t , \quad\quad & \theta^i_{t,\sigma} & =  \ssigma_t / \sn_t \\
\vm_t^i & = \vtheta^i_{t,\mu} [\vh_t^i 1], \quad\quad  & \va_t^i & = \theta^i_{t,\sigma}
\end{alignat}
where $\vm_t^i$ denotes the mean local prediction and $\va_t^i$ denotes the local variance.
The ARU fits local parameters, which can be used to compute the mean
prediction (and variance) much like in the local model in Equation ~\ref{eq:local}.  However, unlike in local models, we do not output these directly as predictions.  Instead, we exploit the availability of multiple time series to further combine these local predictions with global parameters as follows.

\subsection{ARU in Global Model}
\label{sec:aru_in_global_model}
\begin{figure}
    \centering
\vspace*{3mm}
\hspace*{2cm}
\resizebox{7cm}{5.13cm}{
\begin{tikzpicture}[
  hid/.style 2 args={
    rectangle split,
    rectangle split horizontal,
    draw=#2,
    rectangle split parts=#1,
    fill=#2!20,
    outer sep=1mm,
    text width=1cm, 
    text height=1.5mm, 
    text centered}]
  \node[] (x4) at (2*4+4.2, -1) {{{$\vek{x}_4$}}};
  \node[hid={1}{green}] (emb4) at (2*4+4.2, 0) {{{$\vek{x}_4^\text{emb}$}}};
  \node[hid={1}{red}] (g4) at (2*4+1.5+4.2, 0) {{{$\vek{g}_3$}}};
  \node[hid={1}{yellow}] (h4) at (2*4+1+4.2, 1) {{{$\vek{h}_4$}}};
  \node[hid={1}{yellow}] (u4) at (2*4+1+4.2, 5.25) {{{$FF2$}}};
   \node[align=center] (om4) at (2*4+1-0.3+4.2, +6.75) {{{$\hat{\mu}_4$}}};
   \node[align=center] (os4) at (2*4+1+0.3+4.2, +6.75) {{{$\hat{\sigma}_4$}}};
   \draw[->] (x4.north) -> (emb4.south);
   \draw[->] (emb4.north) -> (h4.south);
   \draw[->] (g4.north) -> (h4.south);
   \draw[->] (h4.north) to [bend left] (u4.south);
   \draw[->] (u4.north) -> (om4.south);
   \draw[->] (u4.north) -> (os4.south);
  
   \node[hid={1}{blue}] (arupred4) at (2*5+4.2, 2.25) {ARU\\ Predict};
   \node[hid={1}{blue}] (aruupdate4) at (2*6+4.2, 2.25) {ARU\\ Update};
   \node[] (y4) at (2*6.3+4.2, 1) {{{$\vy_4$}}};
   \node[hid={1}{cyan}] (aruout4) at (2*5+4.2, 3.75) {{{$\vm_{4}, \va_{4}$}}};
   \draw[->] (h4.north) -> (arupred4.south);
   \draw[->] (h4.east) -> (aruupdate4.south);
   \draw[->] (y4.north) -> (aruupdate4.south);
   \draw[->] (arupred4.north) -> (aruout4.south);
   \draw[->] (aruout4.north) -> (u4.south);
   
   \begin{scope}[on background layer]
     \node[fit= (arupred4) (aruupdate4), dashed, draw, inner sep=0.1cm, fill=blue!10] (aru) {};
   \end{scope}
  \node[text width=4cm] at (15+4.2, 2.25) {ARU}; 
 \end{tikzpicture}
}
    \caption{ARU cell combined with the decoder of the global model. Showing only the mean computation here.}
    \label{Fig:aru_in_global_model}
\end{figure}
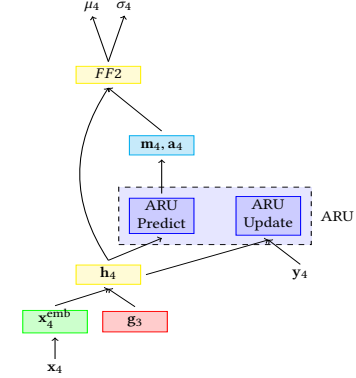

We concatenate the local mean and variance predictions with the final vector $\vh^i_t$ and use a two-layer feed forward network to transform them as follows.
\begin{equation}
\label{eq:aru_in_globalmu}
    \mu^i_t=\theta_\mu (FF2 [\vh^i_t,\vm^i_t,1],~~~~~~~~\sigma^i_t = \log(1+\exp(\theta_\sigma(FF2[\vh^i_t,\va^i_t,1]))
\end{equation}
Figure \ref{Fig:aru_in_global_model} shows how ARU is embedded in the global model for one decoder step. We only show mean computation in the figure. The variance computation can also be handled similarly. The output $\vm_4$ obtained from ARU is combined with hidden layer and passed further to a two layer feed-forward network. The $FF2$ network further evaluates the importance of the global output $\vh^i_t$ and local ARU output $\vm^i_t$ and makes the final predictions. 

\section{Related Work}
Before we present our empirical evaluation we discuss related work that could have served as alternatives to our method of adaptation.  

\paragraph{Generate Local Parameters from RNN State}
Recently \cite{Goel2017} and \cite{Rangapuram2018} propose to use the power of deep learning to directly compute local parameters $\vtheta^i$ from the local state 
of the RNN. 
As a state-of-art representative of this class of methods, we discuss the DeepState method of \cite{Rangapuram2018}. In DeepState the RNN state $\vg^i_t$ computed from global parameters (Eq~\ref{eq:globalmu}) is passed through feed-forward networks to directly output the local parameters  $\theta^i_{\text{tr}}$ and $\theta^i_\mu,
\theta^i_\sigma$, that is,
\begin{eqnarray}
  [\theta^i_{\text{tr}}, \theta^i_\mu,
\theta^i_\sigma] = FF(\vg^i_t; \theta_\text{meta})
\end{eqnarray}
where $\theta_\text{meta}$ are additional global parameters.  The 
computed 
local parameters are used to generate the mean and variance of the output $y$s using a linear state-space model (e.g. Equation~\ref{eq:local}). The global parameters are trained via joint likelihood on all training examples.  Unlike our approach, these methods cannot adapt to new time series since local $y$-s are only transferred via the joint likelihood based training.  In our experiment section we show that these models do not train well unless the length of each time-series is long.

\paragraph{Dedicated Local parameters jointly trained with Global} An approach often used for localization of global models is to dedicate per-locale parameters that play specific role in the global model and are trained jointly end-to-end. For example~\cite{Rei15} uses per-document parameter along with global parameters to develop better embeddings. In our case, we would train time-series specific parameters $\vtheta^i$ along with the rest of the global parameters.  This method is only applicable when all time-series are known during training time.  A downside of this approach is that it cannot be easily evolved when new values are observed for a time series.  Also, each time series has to be long enough for the local parameters to train well. 

\paragraph{Fine-tune Global Parameters for each series}
Another approach is to treat localization as a problem of domain adaptation, for which a huge literature exists~\cite{bendavid06Analysis,BlitzerMP06}.  A well-known solution is to fine-tune the global parameters on labeled data of each time-series using gradient descent on the loss over the limited labeled data for each series~\cite{ravi17}. 
This method would require storage of separate local parameters for each series.  Also, parameter fine-tuning could lead to unpredictable local performance.  This has led to an explosion of meta-learning methods that exploit multiple local datasets during training so as to learn the adaptation process~\cite{Finn2017,mishra2018a}. 
Another method of adaptation is using memory~\cite{Shankar2018, rae2018fast}
that combines parameter fine-tuning with memory-based recall.
Of these a recent state-of-art approach is the SNAIL model that uses a simple method of learned deep self-attention on local data for adaptation.  As a state-of-art representative of learned adaptation methods, we will compare with this method in our experiments.    Our approach is closed-form self-attention that exploits the special regression form of our prediction function. Self-attention based models for time series forecasting are also explored in \cite{QinSCCJC17}.

For theoretical insights on the learning efficiency of global vs local models see \cite{Kuznetsov2019}.



\paragraph{Special RNNs for Time Series}
We designed ARU so that it can be embedded in a global network much like any other RNN.  Recently, \cite{OlivaPS17} and \cite{ZhangLSD18} also propose special RNNs for time-series. Oliva et al. ~\cite{OlivaPS17} proposed an un-gated alternative to LSTMs, called the Statistical Recurrent Unit (SRU). SRUs maintain moving averages of summary statistics in their hidden states and generalize exponential moving average like statistics.  However, they treat the SRU state like 'context' to be used for predictions in place of LSTM states.  Fourier Recurrent Units FRUs~\cite{ZhangLSD18} is another such un-gated time-series unit.  In contrast, we convert the ARU state to local parameters by exploiting the closed-form fit from the the least square sufficient statistics.  \nocite{Li:2010}



\section{Experiments}
We now empirically compare our ARU-based streaming adaptation method to state-of-the-art deep learning based global models as baselines, and to two recent approaches of adapting them to local time-series. In addition to five real-life public time-series datasets, we also present controlled experiments on synthetic datasets to gain insights on different methods of localizing deep global models.  
Our code and experiments can be found at \footnote{\url{https://github.com/pratham16/ARU.git}}.

This section is organized as follows: Section~\ref{sec:data} describes the datasets, Section~\ref{sec:methods} describes the five methods we compare with, Section~\ref{sec:setup} describes detailed setup of the experiments, Section~\ref{sec:synth} presents experiments on the synthetic dataset to qualitatively understand the difference between DeepState and ARU, and Section~\ref{sec:real} presents anecdotes, accuracy, and running times on the real datasets.

\begin{table}[]
    \centering
    \begin{tabular}{l|r|r|r|r|r}
         Dataset & N & T & K & Enc.Len & \#Features \\ \hline
          Rossman &  1115 & 1600 & 16 & 16 & 39 \\
         Walmart &  3331 & 143 & 8 & 8 & 16 \\
         Electricity &  370 & 44000 & 24 & 168 & 5 \\
         Traffic &  963 & 2100 & 24 & 168 & 3 \\
         Parts &  2246 & 52 & 8 & 8 & 1 \\
    \end{tabular}
    \caption{Statistics of the datasets used in our experiments. N is the number of time series, T is the number of values in a series, K is the forecast horizon we used in our experiments, Enc.Len is the length of the encoder in each prediction instance.}
    \label{tab:data}
\end{table}

\subsection{Datasets}
\label{sec:data}
We perform our experiments on five publicly available time-series datasets.  The first two datasets we collected from Kaggle contains rich $\vx$ features at each point and are more aligned with typical retail forecasting setups where the deep models have gained most traction recently.  The remaining three contain only derived time features and are included for comparison with the existing literature.  A summary of the datasets appears in Table~\ref{tab:data}.

\myparagraph{Rossman} 
The Rossmann dataset\footnote{\url{https://www.kaggle.com/c/rossmann-store-sales/data}} contains daily sales history 
of 1115 Rossmann stores. The data is collected between 1st Jan. 2013 to 31st July 2015, i.e., roughly 900 values per series.  Each time series is associated with various co-variate features. 
These
include predetermined store-specific features such as store type and distance to the nearest competing store and time-varying features such as whether the store is running any promotion, and  external features such as  weather conditions, is current day a school holiday or a state holiday, etc.

\myparagraph{Walmart}
The Walmart\footnote{\url{https://www.kaggle.com/c/walmart-recruiting-store-sales-forecasting/data}} dataset maintains weekly sales in each department of 45 walmart stores from Feb 2010 to Oct 2010. Here, (store, department) tuple uniquely identifies each time-series, resulting in 3331 time-series. Each time-series contains roughly 143 values. 
This dataset also provides promotional features and external features such as temperature, unemployment, consumer price index, major holidays in a given week etc.

\myparagraph{Electricity}
The electricity\footnote{https://archive.ics.uci.edu/ml/datasets/ElectricityLoadDiagrams20112014} dataset contains hourly energy consumption of 370 houses (in kW) from 1st Jan. 2011 to 31st Dec. 2015, a total of 44,000 values per time-series.  

\myparagraph{Traffic}
The Traffic\footnote{https://archive.ics.uci.edu/ml/datasets/PEMS-SF} dataset contains hourly occupancy rates of 963 car lanes in San Francisco bay area. The occupancy rates are in the range $[0, 1]$. The data is collected between 1st Jan. 2008 to 30th March 2009, i.e., 10560 values per series.

\myparagraph{Parts}
This dataset \cite{chapados2014effective} contains monthly sales information of 2246 car parts over 52 months.  This is a small dataset compared to others in our collection. 



\paragraph{Features}
For each dataset, we extract time-related features in addition to any features available in the data (e.g. in Rossman and Walmart). 
For Walmart, we use month of the year and week of the year features. For Rossman, we use day of the month and month of the year features. Since Parts is a monthly dataset, we use month of the year feature for Parts. Since Electricity and Traffic are hourly datasets, we use hour of the day and day of the week features. In addition to these, we use month of the year feature for electricity dataset.
We embed categorical features into real vectors  and rescale the real valued attributes between 0 and 1 across the time-series.

\begin{table}[]
    \centering
    \begin{tabular}{|l|r|c|}
        \hline
        Set & \# RNN units & Hidden layer sizes \\ \hline
        {\tt Small} & 8 & 8,6,6 \\
        {\tt Medium} & 16 & 16,15,10 \\
        {\tt Large} & 50 & 32,20,15 \\ \hline
    \end{tabular}
    \caption{Hyperparameter sets}
    \label{tab:hyperparameters}
\end{table}
\begin{figure*}
    \centering
    \includegraphics[width=0.4\textwidth]{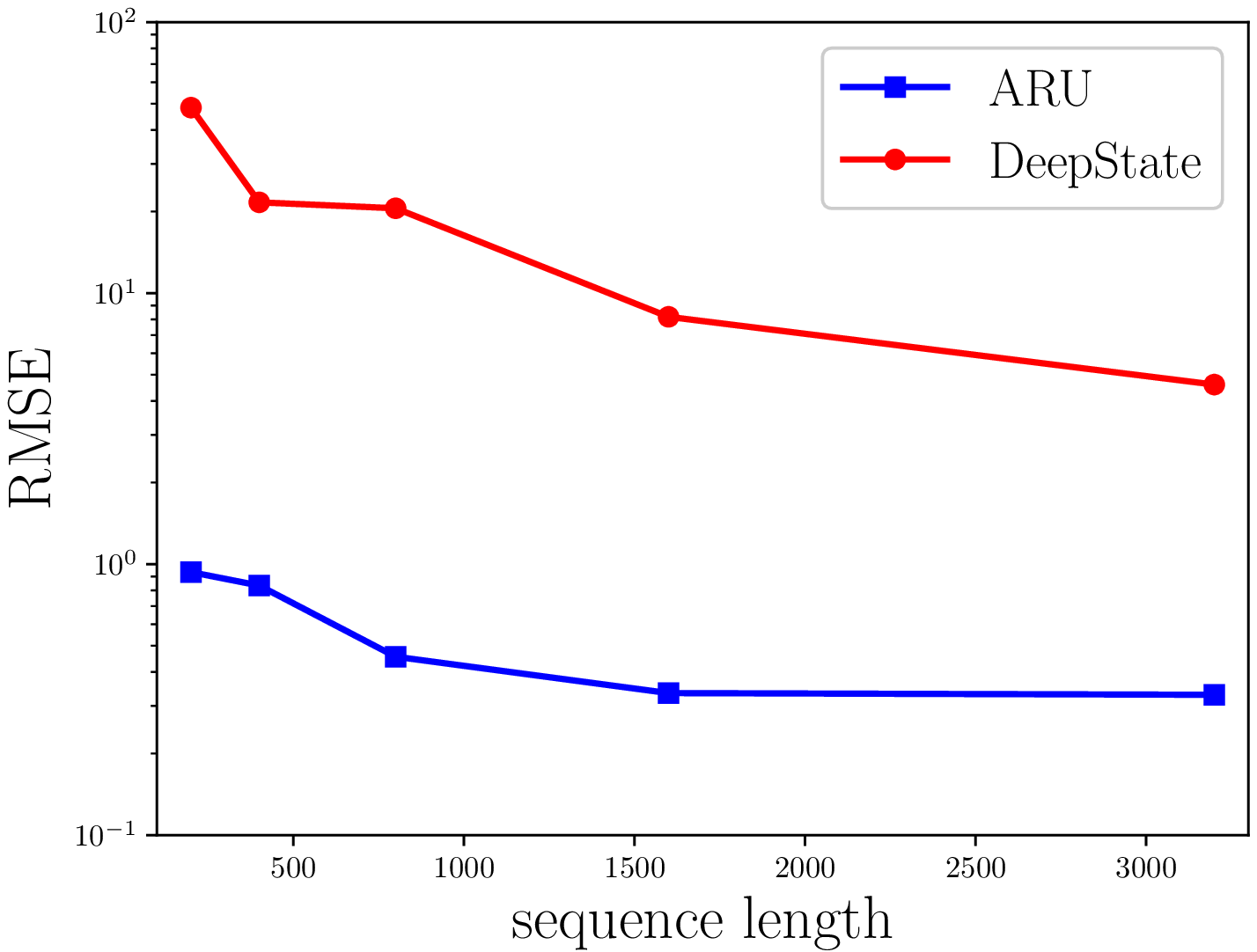}
    \includegraphics[width=0.4\textwidth]{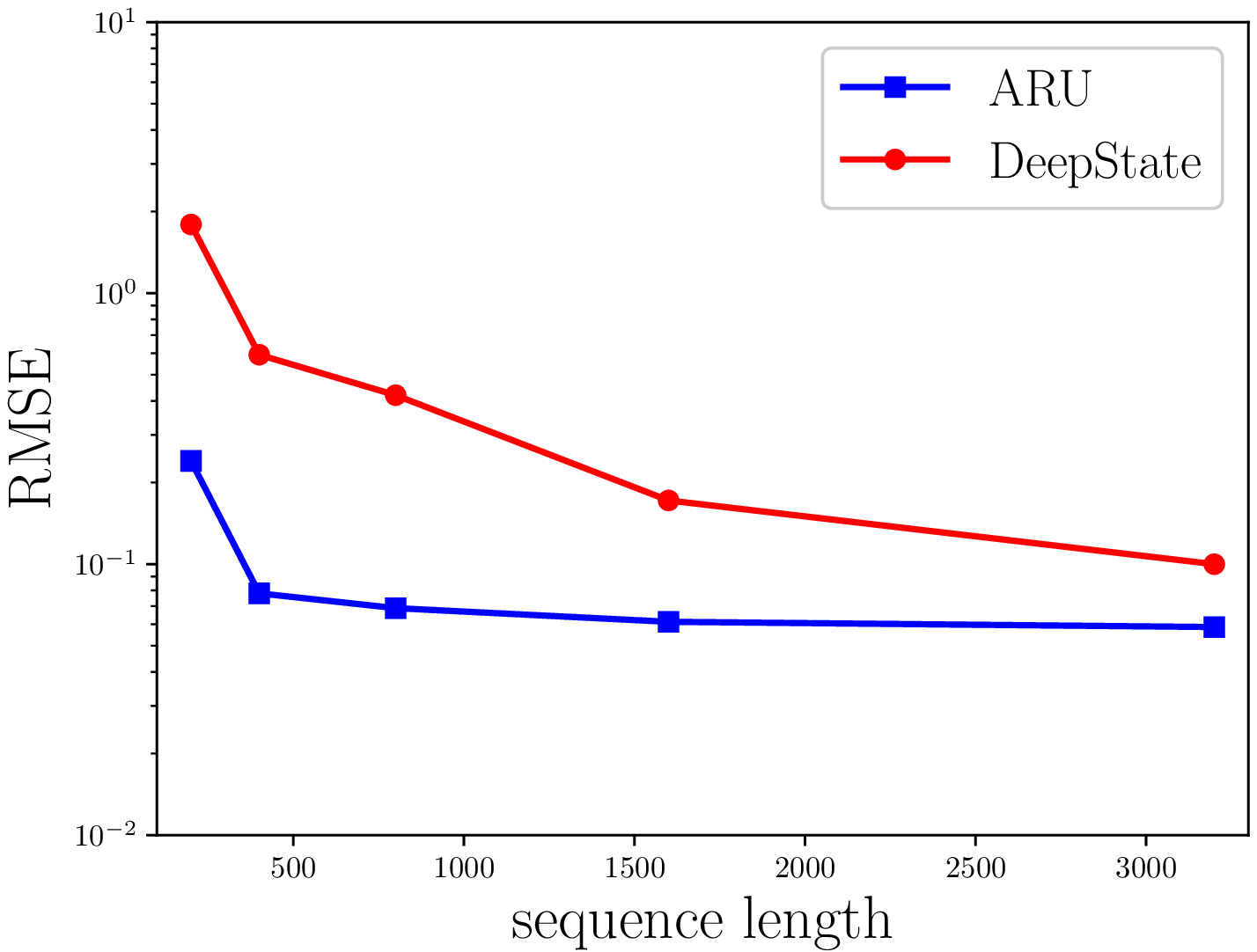}
    \includegraphics[width=0.4\textwidth]{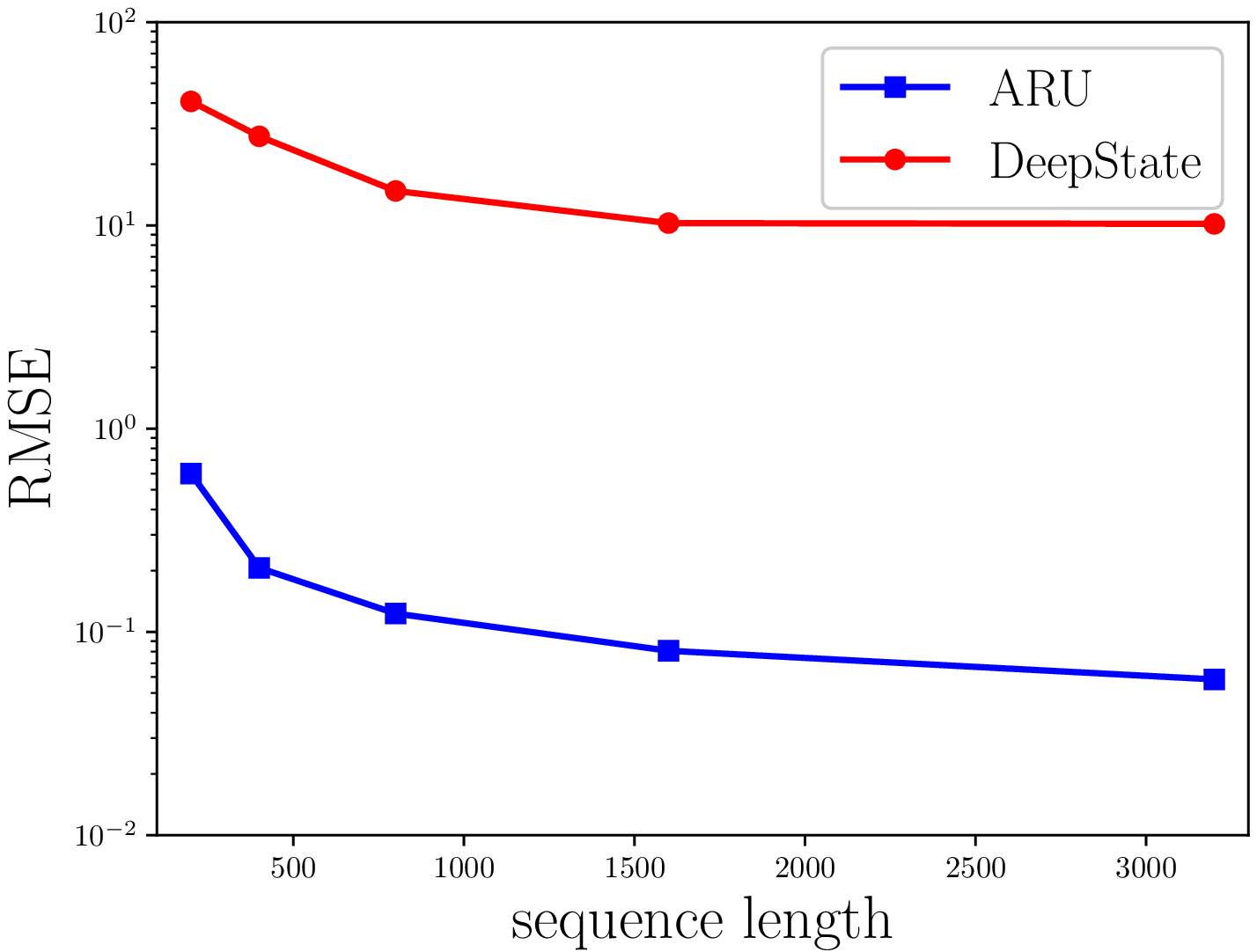}
    \includegraphics[width=0.4\textwidth]{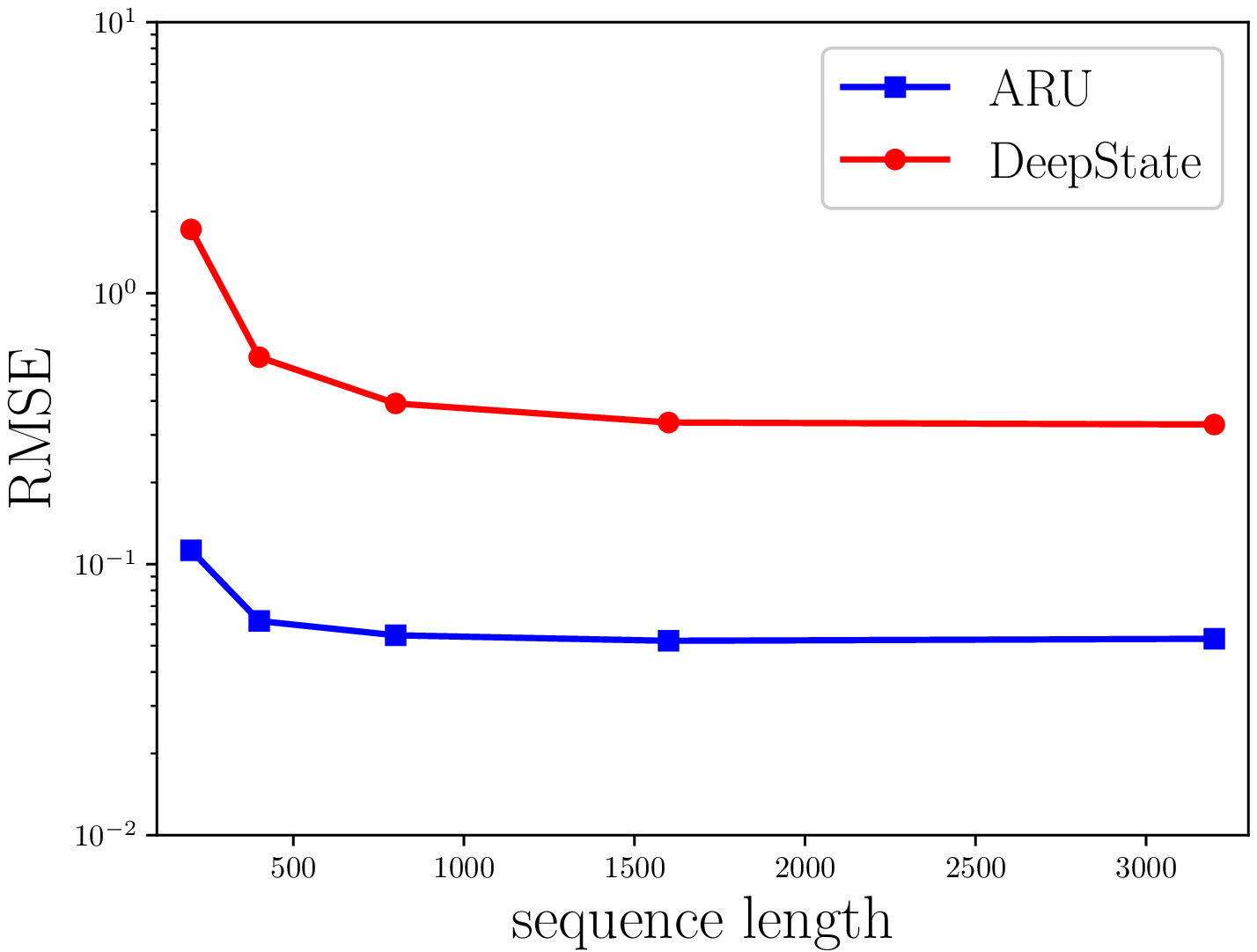}
    \caption{Performance of ARU and DeepState on synthetic data. The columns correspond to different values of $\gamma$ ($\gamma=20$ first column, $\gamma=1$ second column). The first row are with time-series ids and second row are without time-series ids.}
    \label{fig:synth}
\end{figure*}
\subsection{Methods}
\label{sec:methods}
We compare ARU against two state-of-the-art global deep learning models as baselines and two most recent 
local adaptation methods.
We drop comparison with classical local-only models such as ARIMA because the DeepState and DeepAR methods that we compare with have been shown to conclusively surpass them.

\myparagraph{Baseline}
As a baseline we use the globally trained model where the decoder makes independent predictions for each 
of the future $K$ points in time. 
Note that the encoder used to compute the per-series state $\vg^i_T$ continues to be auto-regressive.
Our experiments and \cite{wen2017multi} have found the independent decoder to yield 
better performance 
than an auto-regressive decoder.  This baseline is part of our code base and uses exactly the same setup as ARU.

\myparagraph{DeepAR}
As another global model, we use 
DeepAR, a production quality deep forecasting model that is built-in Amazon's SageMaker machine learning service \footnote{\url{https://docs.aws.amazon.com/sagemaker/latest/dg/deepar.html}}. As described in \cite{FlunkertSG17}, DeepAR is also an encoder-decoder model but their decoder is auto-regressive.  Since the source code is not public, further details of specific features, scaling, etc that they deploy is unknown. To the best of our ability we tried to match the features and experiment setup with our global baseline but we cannot be sure without access to the source code. 

\myparagraph{SNAIL} The SNAIL method proposed in \cite{mishra2018a} is a local adaptation model that uses interleaving of dilated causal convolution and self-attention layers to capture long range dependencies and for localization respectively. We use two convolution layers with dilation rates 2 and 4 respectively, and two self-attention layers. 

\myparagraph{DeepState} This model uses an RNN to predict the parameters of a local state-space model \cite{Rangapuram2018} and is currently the 
best known local adaptation model for time-series forecasting. 
The source code is not publicly available, so we only compare with their published numbers.  



\subsection{Experiment setup}
\label{sec:setup}
We use a single RNN layer in the encoder. The decoder has three ReLU layers on a concatenation of the encoder state $\vg^i_T$ and input features $\vx^i_t$. A skip connection from $\vx^i_{T+k}$ to the second ReLU layer is added. 
The batch size for all experiments is set to 64. Adam optimizer is used with learning rate 0.0001. ARU regularization $\lambda$ and aging vector $\valpha$ are chosen based on validation loss. 
In all our experiments, $\valpha \in \lbrace 1.0, 0.99, 0.95, 0.9 \rbrace^J$. In all experiments, best performing value of $\valpha$ turns out to be $[1.0]$, except for Traffic dataset in Table \ref{tab:deepstate}, where best performing $\valpha = [1.0, 0.95, 0.9]$.  For number of RNN units and number of hidden units, we use three sets of hyper-parameter settings for baseline and ARU -- {\tt Small},  
{\tt Medium}, 
and {\tt Large} as given in Table \ref{tab:hyperparameters}. We use the {\tt Small} configuration for Parts since it contains time-series of length 52. For Rossman and Walmart, we use the 
{\tt Medium} 
configuration. For Electricity and Traffic datasets, we use the {\tt Large} configuration in Table \ref{tab:deepstate} and the 
{\tt Medium} 
configuration in the rest of the experiments. 
%
%
We rescale $y$ values of each time-series with its average $y$ value as follows:
$$ {\overline{y}}^i_t = 1 + \frac{1}{T} \sum_{t'=1}^T  y^i_{t'} $$
This rescaling is also used by \cite{FlunkertSG17}. On the Walmart dataset since each time-series is small, we perform above rescaling on a batch instead of the entire time-series i.e.\ $ {\overline{y}}^i_t = 1 + \frac{1}{E} \sum_{t'=1}^{E} y^i_{t'} $ where $E$ is encoder length.

We also experimented with SRU \cite{OlivaPS17}, which is an un-gated alternative to LSTMs. Since we obtained similar results with SRUs, we omit those numbers from our experiments.

In addition to RMSE, we report Normalized Deviation (ND), which is used in \cite{FlunkertSG17, yu2016temporal} as an evaluation metric. It is defined as:
$$ \text{ND}(y^i_t, \hat{y}^i_t) = \bigg( \sum_{i=1}^N \sum_{t=T+1}^{T+K} |y^i_t - \hat{y}^i_t| \bigg) \bigg/ \bigg( \sum_{i=1}^N \sum_{t=T+1}^{T+K} |y^i_t| \bigg) $$


\begin{figure*}
    \centering
    \includegraphics[width=0.45\textwidth]{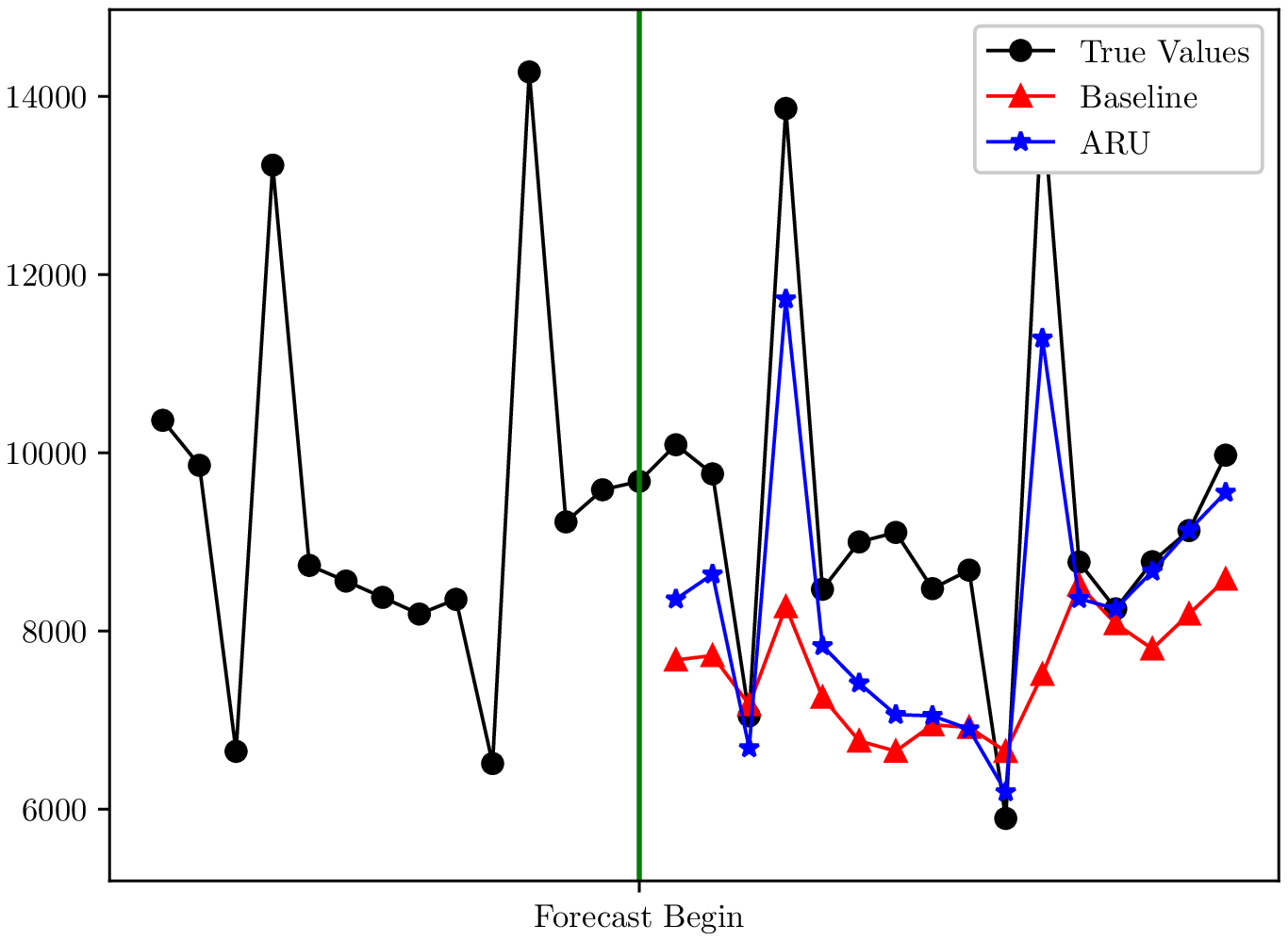}
    \includegraphics[width=0.45\textwidth]{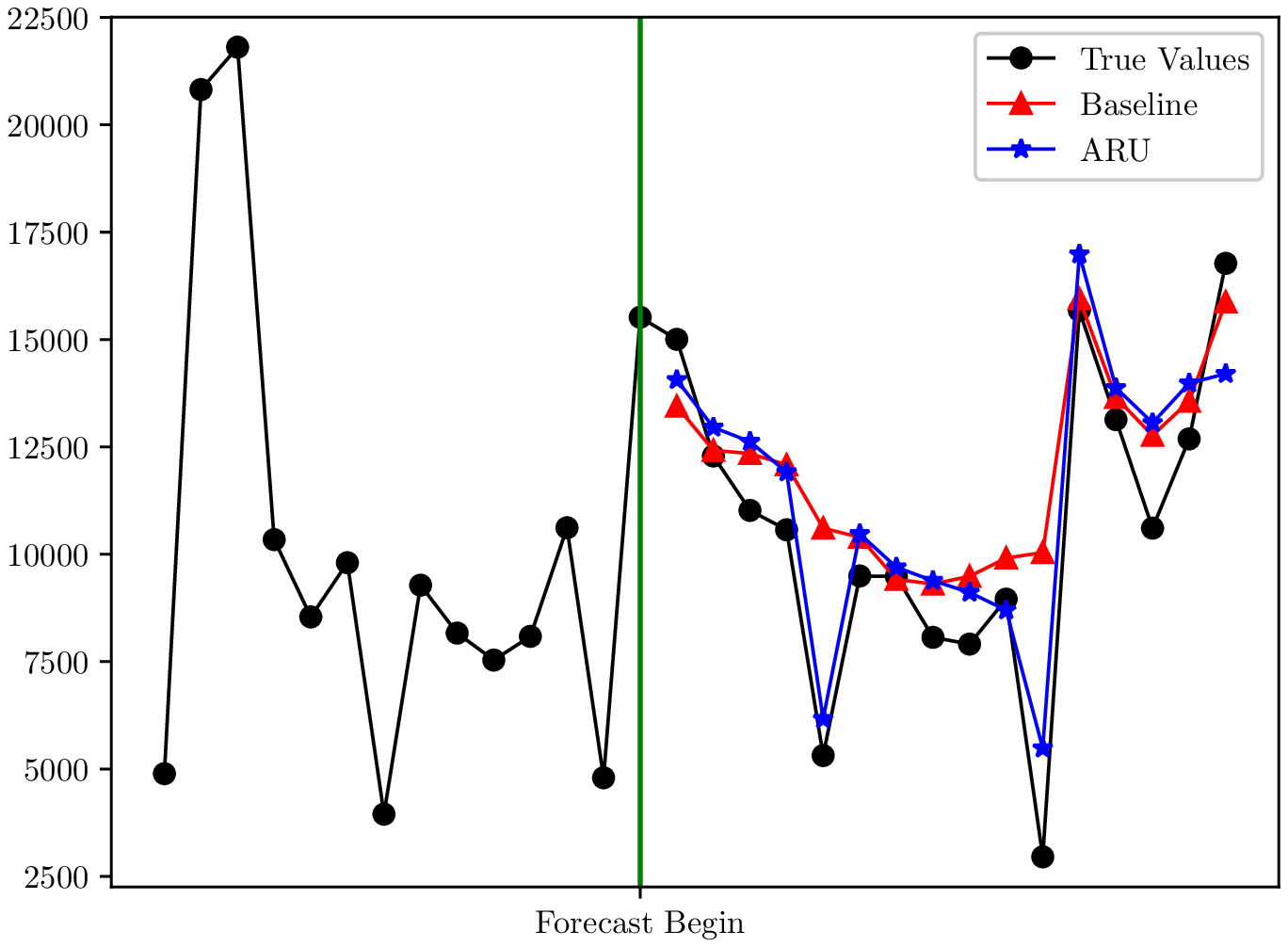}

    \includegraphics[width=0.45\textwidth]{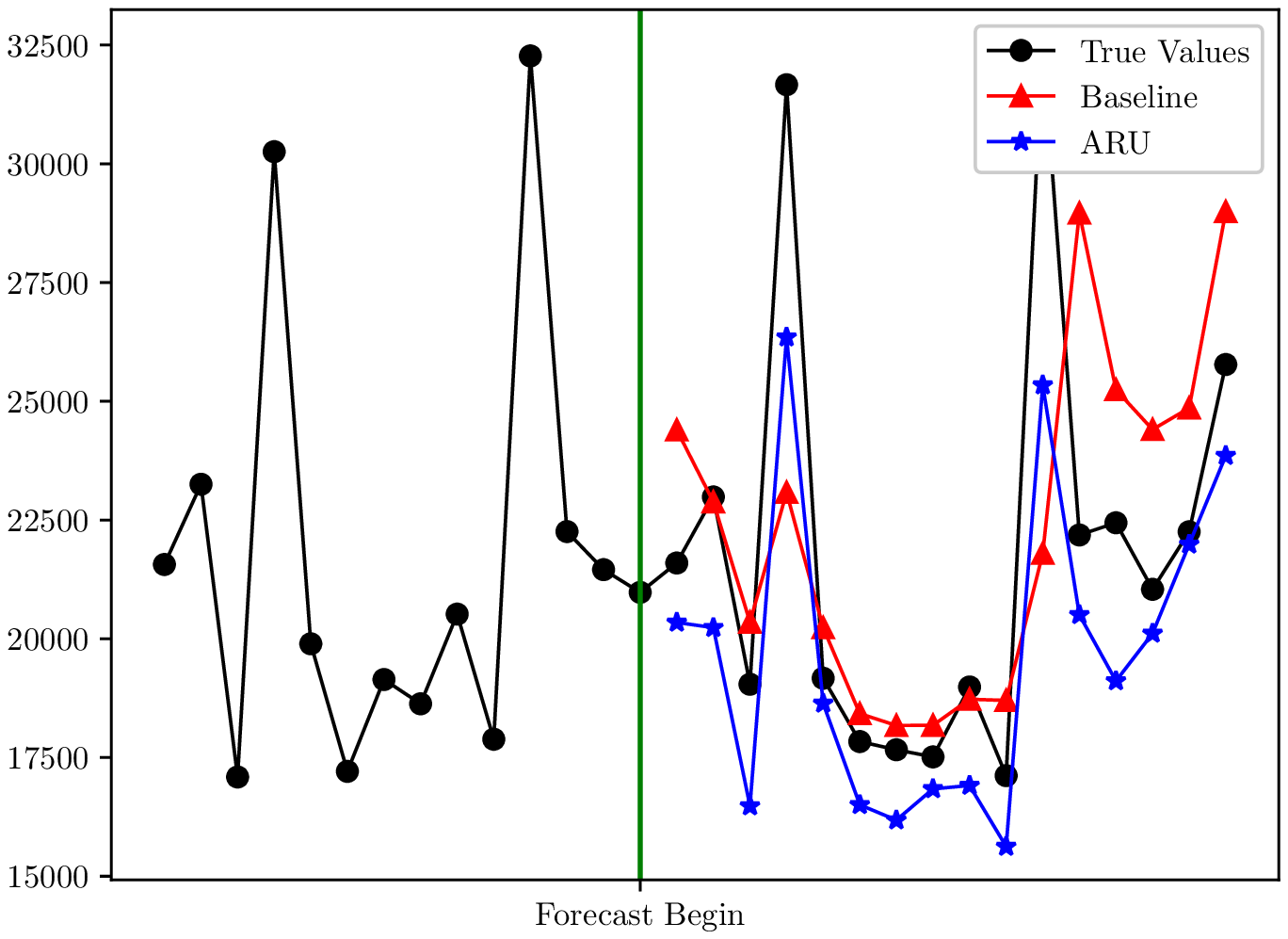}
    \includegraphics[width=0.45\textwidth]{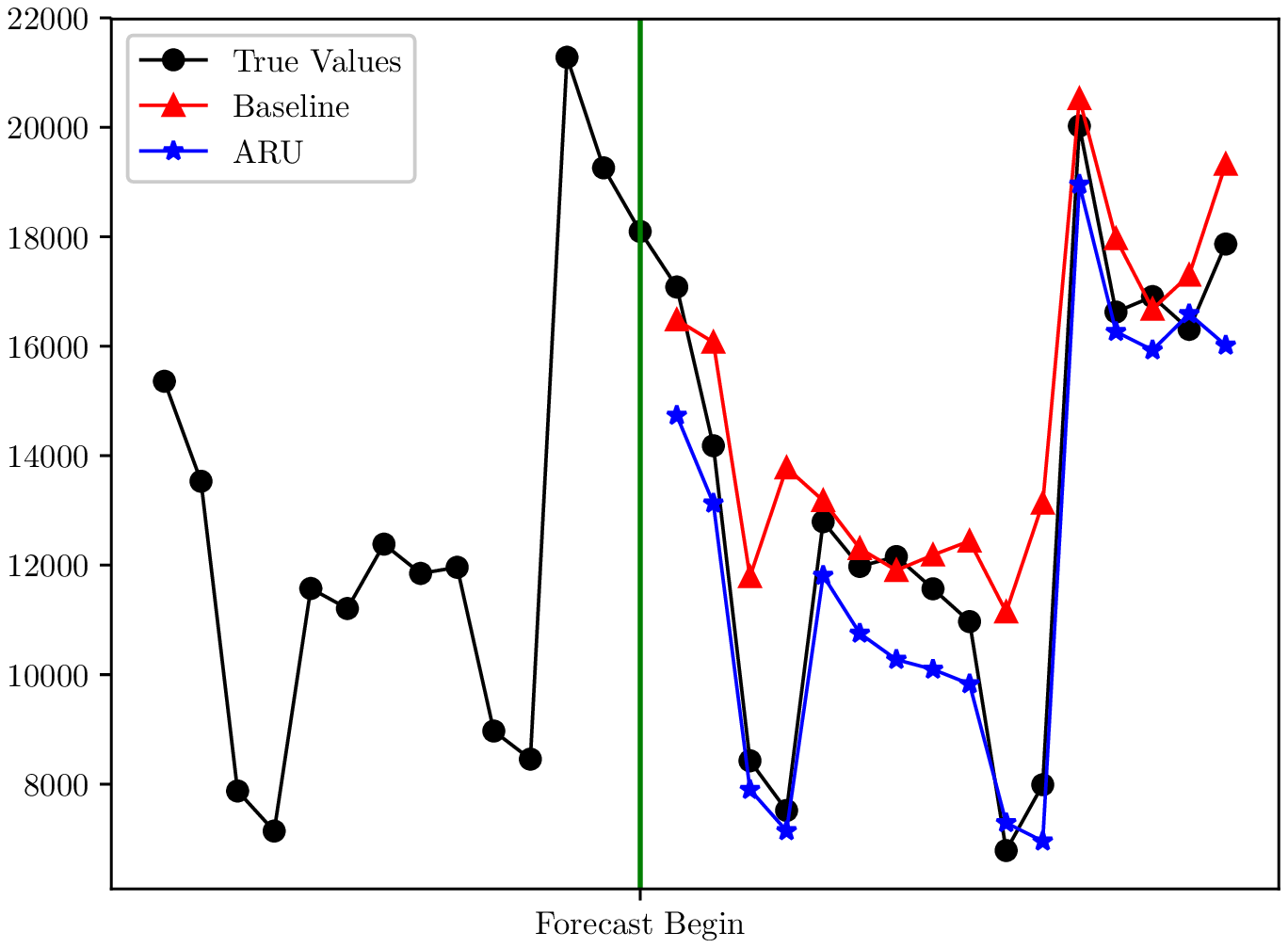}

    \caption{Examples of four time series from the Rossman dataset showing how well our ARU is able to track the local patterns of each series compared to the global baseline.}
    \label{Fig:example_rossman}
\end{figure*}
\subsection{Qualitative Comparison on Synthetic Data}
\label{sec:synth}
In order to get an insight into 
the working of ARU vis-a-vis the most recent DeepState method of local adaptation, we first present a qualitative comparison using a very simple synthetic dataset.

We generate 10 synthetic time-series of hourly data. The $y$ values for these time-series are obtained as a linear combination of $x$ features and local parameters $\vtheta_\mu^i$, which are uniformly sampled in the range $[-\gamma,\gamma]$. We use only hour of day and day of week as $\vx$ features and generate time-series of different lengths 
by sampling $y \sim \vtheta_\mu^i [\vx_t~1] + \epsilon$ where $\epsilon \sim \mathcal{N}(0, 1)$ is a Gaussian noise with unit variance.
We compute RMSE for predictions on last 24 hour forecasts in each time-series. This non-auto-regressive model might look too simplistic but it helps us bring out the essence of different approaches.  
Also, after conditioning on the encoder-state, we found non-auto-regressive decoder to be superior to the auto-regressive decoder even on real-life datasets. 
In this simple case, the state-space model would need to learn to generate  $\vtheta_\mu^i$ from its RNN state $\vg^i_t$ in order to match the generative process exactly.  In contrast, the ARU would compute
$\vtheta_\mu^i$ from the data whose sufficient statistics is summarized in the ARU state. The global network just needs to learn to provide the right input to the ARU cell and also correctly integrate its output with the global output.  
We used the {\tt Medium} setting (Table~\ref{tab:hyperparameters}) for the global network without any special customization for the synthetic experiments. Thus, the global network needs to learn the correct meta-learning strategy for both methods of adaptation.
When id $i$ is part of $\vx_t$, in theory, a powerful enough network can memorize each set of the 10 local parameter in its feed-forward layers, making them both identical.  We therefore compare these models under two settings: one where each time-series $i$ feeds its unique id $i$ to the network, and second where it does not.


%
%
We compare the two methods against increasing length of time-series in Figure~\ref{fig:synth} for two different values of $\gamma$ (20 and 1) both with and without time-id.  When $\gamma$ is large (first column) the local parameters differ a lot from one time series to the next and the DeepState method of computing them from the RNN state is not at all effective.  With increasing sequence length, errors of both methods reduce but ARU converges much faster to error values less then 0.1. When local time-series are more similar to each other ($\gamma=1$), DeepState starts matching ARU only after seeing many samples per time-series ($\gamma=1$, with time-series id).  Also, parameter generation methods like DeepState also crucially exploit the time-series specific identifiers  to generate local parameters (compare red plots across the two rows). This implies that they memorize local parameters of the finite number of training time-series, and cannot perform adaptation of a series not seen during training. Also, they cannot exploit streaming availability of labeled data after training.



\begin{table}[ht]
    \centering
    \begin{tabular}{|l|l|r|r|} \hline
    Dataset & Method & Normalized & RMSE \\ 
             &       & Deviation (ND)   &  \\ \hline
        Rossman  & Baseline  & 0.0987  & 1020.7 \\ 
             & DeepAR & 0.2273 & 2197.4 \\ 
             & SNAIL & 0.0963 & 993.4 \\ 
              & \ModifiedARU & {\bf 0.0851} & {\bf 908.6} \\ \hline
    Walmart  & Baseline & 0.1187 & 3585.8 \\
             & DeepAR & 0.2215 & 5894.9 \\
             & SNAIL & 0.1082 & 3465.0 \\
             & \ModifiedARU & {\bf 0.1058} & {\bf 3413.3} \\ \hline
       
    Electricity  & Baseline  & 0.1264 & 377.3 \\ 
             & DeepAR & 0.1838 & 530.9 \\ 
             & SNAIL & 0.1307 & 363.2 \\ 
              & \ModifiedARU & {\bf 0.1260} & {\bf 344.6} \\ \hline
              
    Traffic  & Baseline  & 0.1991 & 0.0374 \\ 
             & DeepAR & {\bf 0.1894} & 0.0367 \\ 
             & SNAIL & 0.2187 & 0.0380 \\ 
              & \ModifiedARU & 0.1964 & 0.0372 \\ \hline

      Parts  & Baseline  & 1.5763 & 1.300 \\ 
             & DeepAR & {\bf 1.4440} & {\bf 1.268} \\ 
             & SNAIL & 1.4910 & 1.276 \\ 
              & \ModifiedARU & 1.5409 & 1.285 \\ \hline              

    \end{tabular}
    \caption{Performance of different methods on the datasets in the fixed mode.}
    \label{tab:overall}
\end{table}

\subsection{Comparisons on Real Data}
\label{sec:real}
We start by showing some anecdotes, then compare different methods on prediction error in two different settings, and finally compare them on running time.

\subsubsection{Anecdotes} In Figure \ref{Fig:example_rossman}, we show how ARU enables a global model to adapt to the 
characteristics 
of each time series from the Rossman dataset.  The vertical line denotes the prediction horizon.  The ARU model is able to match the peaks of the data much better than the global model.  The location and pattern of peaks in the actual data varies a lot across the four time-series and a single global model is unlikely to be able to capture them.  The ARU by intervening at the last layer, with local sufficient statistics, is able to learn the relation between the inputs and peaks 
more accurately. 
For example, look at the two peaks after the green line in the first plot.  For the time-series in the second row, first column we see how the global model's peak prediction is offset by one 
step 
whereas ARU aligns perfectly with the true $y$s.  In the last time-series (second row, second column), we see how nicely the gradual decline is matched by ARU, whereas the global baseline is needlessly jittery.

\subsubsection{Quantitative comparison on prediction error}
Next we move to a quantitative comparison.  We first compare in a fixed horizon setting where we predict for a fixed horizon 
$T+1,\ldots,T+K$
that is known during training ($T$ and $K$ values summarized in Table~\ref{tab:data}). 
In Table~\ref{tab:overall} we compare different methods for a fixed horizon on the five datasets on four methods. 
The DeepState method is compared separately since 
we could not get
access to the DeepState code base.
We observe that on feature-rich datasets like Rossman and Walmart ARU is significantly better than all existing methods.  
Another property of ARU is that even when it does not improve beyond the baseline it is not much worse. This is because we meta-learn how to combine ARU's local predictions with the global predictions.

We next compare different methods 
in 
a streaming setting 
where we provide the true $E$ (encoder length) forecasts after the training time period $T$, and ask for prediction for the next $K$ time periods.
We do not retrain model parameters to simulate a true online deployment setting.  For the electricity and traffic datasets, we use the rolling-window forecast setting used in \cite{Rangapuram2018} and \cite{yu2016temporal}. Hourly values of 7 days are input and we make prediction for the next day. Then the forecast window is shifted by one day and we repeat this for the next 7 days.  Since the frequency of the Walmart dataset is weekly, we make rolling-window forecasts for 16 weeks, with window size 8. For the Rossman data, forecast window size is 16 days, and we make forecasts for 32 days.

\begin{table}[ht]
    \centering
    \begin{tabular}{|l|l|r|r|} \hline
      Dataset & Method & Normalized & RMSE \\ 
             &       & Deviation (ND)   &  \\ \hline
    Rossman  & Baseline  & 0.094 & 983.2 \\ 
             & DeepAR & 0.245 & 2389.0 \\ 
             & SNAIL & 0.093 & 972.6 \\ 
             & \ModifiedARU & {\bf 0.089} & {\bf 934.9} \\ \hline
 
    Walmart 
             & Baseline  & 0.137 & 4548.6 \\
             & DeepAR & 0.233 & 6704.0 \\ 
             & SNAIL & 0.144 & 4690.5 \\ 
             & \ModifiedARU & {\bf 0.114} & {\bf 3938.6} \\ \hline
              
    Electricity  
             & Baseline & 0.136 & 416.7 \\
            & DeepAR & 0.172 & 544.0 \\ 
             & SNAIL & 0.135 & 400.6 \\ 
             & \ModifiedARU & {\bf 0.127} & {\bf 396.4} \\ \hline
              
    Traffic
             & Baseline & 0.170 & 0.0224 \\
             & DeepAR & {\bf 0.145} & {\bf 0.0216} \\
             & SNAIL & 0.165 & 0.0227 \\ 
             & ARU & 0.161 & 0.0220 \\ 
   \hline
    \end{tabular}
    \caption{Performance of different methods in the streaming mode.}
    \label{tab:stream}
\end{table}
In Table~\ref{tab:stream}, we summarize the results on four datasets. The car parts dataset was dropped since each time-series is too small for streaming.  We observe that even in this mode, ARU provides significant gains over all existing methods. A close second is the SNAIL method of adaptation but we will see later that SNAIL incurs significant runtime overheads.

\begin{table}[]
    \centering
    \begin{tabular}{|l|l|l|l|l|} \hline
         Dataset &  Setting & DeepAR & DeepState & ARU \\
         &   Enc Len-$K$ &  & & \\ 
         \hline
         Parts & 12-12 & 1.245 & 1.470 & 1.318 \\ 
         Electricity & 336-168 & 0.199 & 0.087 & 0.116 \\
         Traffic  & 336-168 & 0.148 & 0.168 & 0.159 \\
         Tourism-monthly & 48-24 & 0.107 & 0.138 & 0.096 \\ \hline
    \end{tabular}
    \caption{ND loss for datasets on which DeepState reported results. The setting denotes the respective encoder and decoder lengths (prediction horizon).}
    \label{tab:deepstate}
\end{table}
\subsubsection{Comparison with DeepState}
Since the DeepState code is not publicly available, we  compare ARU with DeepState on exactly those dataset and settings for which we were able to approximately match the publicly available DeepAR numbers in the DeepState paper.  Here we also include a sixth dataset Tourism-monthly since it was used in 
the DeepState paper. 
This is a small dataset of 366 time-series containing monthly tourism demand for approximately 15 years \cite{athanasopoulos2011tourism}. In this dataset, the start timestamp as well as length of each time-series are different. Since this is a monthly dataset, we use month of the year feature. 
We observe that ARU is better than DeepState in 3 out of 4 datasets.  Both DeepAR and DeepState models use a time-series id, whereas we do not. 
Perhaps these datasets do not show too much local variation, making it possible for large deep models to memorize local patterns in its parameters as we showed in our synthetic experiments.




\subsubsection{ARU integration}
Another way in which we differ from DeepState is that instead of directly 
yielding 
the local predictions, we integrate them with global predictions using more learned global parameters.  This allows us to safeguard against the high variance in local predictions and to meta-learn the best fall-back strategy.  In Table~\ref{tab:ARU-Direct} we compare ARU with ARU-direct whose local predictions are output directly.  We observe that ARU-Direct is much worse than ARU and also worse than the Baseline. 
\begin{table}[]
    \centering
    \begin{tabular}{|l|l|r|r|} \hline
    Dataset & Method & Normalized & RMSE \\ 
             &       & Deviation (ND)   &  \\ \hline
        Rossman  & Baseline  & 0.0987  & 1020.7 \\
              & \ModifiedARU & 0.0851 & 908.6 \\ 
              & ARU-Direct & 0.1358 & 1835.68 \\ \hline
              
    Walmart  & Baseline & 0.1187 & 3585.8 \\
             & \ModifiedARU & 0.1058 & 3413.3 \\ 
             & ARU-Direct & 0.1271 & 4041.88 \\\hline
       
    Electricity  & Baseline  & 0.1264 & 377.3 \\
              & \ModifiedARU & 0.1260 & 344.6 \\ 
              & ARU-Direct & 0.1389 & 398.11 \\ \hline
              
    Traffic  & Baseline  & 0.1991 & 0.0374 \\
              & \ModifiedARU & 0.1964 & 0.0372 \\ 
              & ARU-Direct & 0.2195 & 0.0379 \\ \hline

      Parts  & Baseline  & 1.5763 & 1.300 \\
              & \ModifiedARU & 1.5409 & 1.285 \\ 
              & ARU-Direct & 1.6555 & 1.5332 \\ \hline
    \end{tabular}
    \caption{Comparison of performance of ARU-Direct method with Baseline and \ModifiedARU}
    \label{tab:ARU-Direct}
\end{table}

\begin{table}[]
\setlength\tabcolsep{10.0pt}
    \centering
    \begin{tabular}{|l|l|l|l|} \hline
         Dataset & Baseline & ARU & SNAIL \\ \hline
         Electricity & 1.0458 & 1.1788 & 3.0754 \\
         Traffic & 2.0740 & 2.3383 & 5.7673 \\
         Walmart & 0.7034 & 0.9434 & 1.2693 \\
         Rossman & 0.4379 & 0.6717 & 2.2837 \\ \hline
    \end{tabular}
    \caption{Inference time (in seconds)}
    \label{tab:running_time}
\end{table}
\subsubsection{Running time comparison}
In Table \ref{tab:running_time}, we compare running times of different models of inference on the test data. The running time of SNAIL is larger, up to a factor of four higher than the baseline, due to its costly self-attention mechanism. In contrast, the running time of ARU is at most 
1.5 times more than the baseline.
Also, since ARU stores a fixed sized state per time-series, its storage overhead is also very small. 

These experiments demonstrate that ARU is an effective light-weight method of streaming adaptation that can be easily plugged into any existing global model.
\section{Conclusion}
In this paper we presented ARU, an adaptive recurrent unit that provides streaming local adaptation of globally 
trained 
deep time-series models. The core principle in ARU is simple --- exploit the kernel trick to obtain a closed-form optimal solution to linear least square regression loss.  While this trick is classically known, our contribution is in recognizing the imminently practical use of this trick in modern deep network settings and designing a method of effectively integrating global-local patterns.  Unlike existing self-attention or memory-augmented models with linearly increasing memory requirements, ARU allocates only constant sized state per time-series.  Unlike methods that learn to generate local parameters, ARU can adapt to streaming data.  Experiments on five real-life and synthetic datasets establish that ARU is an effective light-weight method of streaming adaptation of global deep forecasting models.
In future, we plan to apply the ARU method of adaptation to linear layers in other parts of a deep network. 

\noindent
\paragraph{Acknowledgements: }We thank Flipkart for sponsoring the project, Srayanta Mukherjee and P Kompalli for getting us interested in the problem, and M Prashanth for performing initial  experiments. 
\bibliographystyle{ACM-Reference-Format}
\bibliography{ML,mining2}


\begin{thebibliography}{31}


\ifx \showCODEN    \undefined \def \showCODEN     #1{\unskip}     \fi
\ifx \showDOI      \undefined \def \showDOI       #1{#1}\fi
\ifx \showISBNx    \undefined \def \showISBNx     #1{\unskip}     \fi
\ifx \showISBNxiii \undefined \def \showISBNxiii  #1{\unskip}     \fi
\ifx \showISSN     \undefined \def \showISSN      #1{\unskip}     \fi
\ifx \showLCCN     \undefined \def \showLCCN      #1{\unskip}     \fi
\ifx \shownote     \undefined \def \shownote      #1{#1}          \fi
\ifx \showarticletitle \undefined \def \showarticletitle #1{#1}   \fi
\ifx \showURL      \undefined \def \showURL       {\relax}        \fi
\providecommand\bibfield[2]{#2}
\providecommand\bibinfo[2]{#2}
\providecommand\natexlab[1]{#1}
\providecommand\showeprint[2][]{arXiv:#2}

\bibitem[\protect\citeauthoryear{Ara\'{u}jo, Ribeiro, and Faloutsos}{Ara\'{u}jo
  et~al\mbox{.}}{2018}]%
        {Araujo:2018}
\bibfield{author}{\bibinfo{person}{Miguel Ara\'{u}jo}, \bibinfo{person}{Pedro
  Ribeiro}, {and} \bibinfo{person}{Christos Faloutsos}.}
  \bibinfo{year}{2018}\natexlab{}.
\newblock \showarticletitle{Tensorcast: Forecasting Time-evolving Networks with
  Contextual Information}. In \bibinfo{booktitle}{\emph{Proceedings of the 27th
  International Joint Conference on Artificial Intelligence}}
  \emph{(\bibinfo{series}{IJCAI'18})}.
\newblock


\bibitem[\protect\citeauthoryear{Athanasopoulos, Hyndman, Song, and
  Wu}{Athanasopoulos et~al\mbox{.}}{2011}]%
        {athanasopoulos2011tourism}
\bibfield{author}{\bibinfo{person}{George Athanasopoulos},
  \bibinfo{person}{Rob~J Hyndman}, \bibinfo{person}{Haiyan Song}, {and}
  \bibinfo{person}{Doris~C Wu}.} \bibinfo{year}{2011}\natexlab{}.
\newblock \showarticletitle{The tourism forecasting competition}.
\newblock \bibinfo{journal}{\emph{International Journal of Forecasting}}
  \bibinfo{volume}{27}, \bibinfo{number}{3} (\bibinfo{year}{2011}),
  \bibinfo{pages}{822--844}.
\newblock


\bibitem[\protect\citeauthoryear{Ben-David, Blitzer, Crammer, and
  Pereira}{Ben-David et~al\mbox{.}}{2007}]%
        {bendavid06Analysis}
\bibfield{author}{\bibinfo{person}{Shai Ben-David}, \bibinfo{person}{John
  Blitzer}, \bibinfo{person}{Koby Crammer}, {and} \bibinfo{person}{Fernando
  Pereira}.} \bibinfo{year}{2007}\natexlab{}.
\newblock \showarticletitle{Analysis of Representations for Domain Adaptation}.
  In \bibinfo{booktitle}{\emph{Advances in Neural Information Processing
  Systems 20}}. \bibinfo{publisher}{MIT Press}, \bibinfo{address}{Cambridge,
  MA}.
\newblock


\bibitem[\protect\citeauthoryear{Blitzer, McDonald, and Pereira}{Blitzer
  et~al\mbox{.}}{2006}]%
        {BlitzerMP06}
\bibfield{author}{\bibinfo{person}{John Blitzer}, \bibinfo{person}{Ryan
  McDonald}, {and} \bibinfo{person}{Fernando Pereira}.}
  \bibinfo{year}{2006}\natexlab{}.
\newblock \showarticletitle{Domain adaptation with structural correspondence
  learning}. In \bibinfo{booktitle}{\emph{Proceedings of the 2006 conference on
  empirical methods in natural language processing}}. Association for
  Computational Linguistics, \bibinfo{pages}{120--128}.
\newblock


\bibitem[\protect\citeauthoryear{Box and Cox}{Box and Cox}{1964}]%
        {box64}
\bibfield{author}{\bibinfo{person}{G.E.P Box} {and} \bibinfo{person}{D.R.
  Cox}.} \bibinfo{year}{1964}\natexlab{}.
\newblock \showarticletitle{An analysis of transformations}.
\newblock \bibinfo{journal}{\emph{Journal of Royal Statistical Society. Series
  B (Methodological)}} \bibinfo{volume}{26}, \bibinfo{number}{2}
  (\bibinfo{year}{1964}), \bibinfo{pages}{211--252}.
\newblock


\bibitem[\protect\citeauthoryear{Box and Jenkins}{Box and Jenkins}{1968}]%
        {box68}
\bibfield{author}{\bibinfo{person}{G.E.P Box} {and} \bibinfo{person}{Gwilym~M.
  Jenkins}.} \bibinfo{year}{1968}\natexlab{}.
\newblock \showarticletitle{Some recent advances in forecasting and control}.
\newblock \bibinfo{journal}{\emph{Journal of Royal Statistical Society. Series
  C (Applied Statistics)}} \bibinfo{volume}{17}, \bibinfo{number}{2}
  (\bibinfo{year}{1968}), \bibinfo{pages}{91--109}.
\newblock


\bibitem[\protect\citeauthoryear{Chapados}{Chapados}{2014}]%
        {chapados2014effective}
\bibfield{author}{\bibinfo{person}{Nicolas Chapados}.}
  \bibinfo{year}{2014}\natexlab{}.
\newblock \showarticletitle{Effective Bayesian modeling of groups of related
  count time series}.
\newblock \bibinfo{journal}{\emph{arXiv preprint arXiv:1405.3738}}
  (\bibinfo{year}{2014}).
\newblock


\bibitem[\protect\citeauthoryear{Faloutsos, Gasthaus, Januschowski, and
  Wang}{Faloutsos et~al\mbox{.}}{2018}]%
        {Faloutsos2018}
\bibfield{author}{\bibinfo{person}{Christos Faloutsos}, \bibinfo{person}{Jan
  Gasthaus}, \bibinfo{person}{Tim Januschowski}, {and} \bibinfo{person}{Yuyang
  Wang}.} \bibinfo{year}{2018}\natexlab{}.
\newblock \showarticletitle{Forecasting Big Time Series: Old and New}.
\newblock \bibinfo{journal}{\emph{Proc. VLDB Endow.}} \bibinfo{volume}{11},
  \bibinfo{number}{12} (\bibinfo{date}{Aug.} \bibinfo{year}{2018}),
  \bibinfo{pages}{2102--2105}.
\newblock
\showISSN{2150-8097}


\bibitem[\protect\citeauthoryear{Ferreira, Lee, and Simchi-Levi}{Ferreira
  et~al\mbox{.}}{2015}]%
        {johnson15}
\bibfield{author}{\bibinfo{person}{Kris~Johnson Ferreira},
  \bibinfo{person}{Bing Hong~Alex Lee}, {and} \bibinfo{person}{David
  Simchi-Levi}.} \bibinfo{year}{2015}\natexlab{}.
\newblock \showarticletitle{Analytics for and online retailer: Demand
  forecasting and price optimization}.
\newblock \bibinfo{journal}{\emph{Manufacturing and Service Operations
  Management}} \bibinfo{volume}{18}, \bibinfo{number}{1}
  (\bibinfo{year}{2015}), \bibinfo{pages}{69--88}.
\newblock


\bibitem[\protect\citeauthoryear{Finn, Abbeel, and Levine}{Finn
  et~al\mbox{.}}{2017}]%
        {Finn2017}
\bibfield{author}{\bibinfo{person}{Chelsea Finn}, \bibinfo{person}{Pieter
  Abbeel}, {and} \bibinfo{person}{Sergey Levine}.}
  \bibinfo{year}{2017}\natexlab{}.
\newblock \showarticletitle{Model-Agnostic Meta-Learning for Fast Adaptation of
  Deep Networks}. In \bibinfo{booktitle}{\emph{Proceedings of the 34th
  International Conference on Machine Learning}}. \bibinfo{pages}{1126--1135}.
\newblock


\bibitem[\protect\citeauthoryear{Flunkert, Salinas, and Gasthaus}{Flunkert
  et~al\mbox{.}}{2017}]%
        {FlunkertSG17}
\bibfield{author}{\bibinfo{person}{Valentin Flunkert}, \bibinfo{person}{David
  Salinas}, {and} \bibinfo{person}{Jan Gasthaus}.}
  \bibinfo{year}{2017}\natexlab{}.
\newblock \showarticletitle{DeepAR: Probabilistic Forecasting with
  Autoregressive Recurrent Networks}.
\newblock \bibinfo{journal}{\emph{CoRR}}  \bibinfo{volume}{abs/1704.04110}
  (\bibinfo{year}{2017}).
\newblock


\bibitem[\protect\citeauthoryear{Goel, Melnyk, and Banerjee}{Goel
  et~al\mbox{.}}{2017}]%
        {Goel2017}
\bibfield{author}{\bibinfo{person}{Hardik Goel}, \bibinfo{person}{Igor Melnyk},
  {and} \bibinfo{person}{Arindam Banerjee}.} \bibinfo{year}{2017}\natexlab{}.
\newblock \showarticletitle{{R2N2:} Residual Recurrent Neural Networks for
  Multivariate Time Series Forecasting}.
\newblock \bibinfo{journal}{\emph{CoRR}}  \bibinfo{volume}{abs/1709.03159}
  (\bibinfo{year}{2017}).
\newblock


\bibitem[\protect\citeauthoryear{Hyndman, Koehler, Ord, and Snyder}{Hyndman
  et~al\mbox{.}}{2008}]%
        {hyndman08}
\bibfield{author}{\bibinfo{person}{R. Hyndman}, \bibinfo{person}{A.B. Koehler},
  \bibinfo{person}{J.K. Ord}, {and} \bibinfo{person}{R.D. Snyder}.}
  \bibinfo{year}{2008}\natexlab{}.
\newblock \bibinfo{booktitle}{\emph{Forecasting with exponential smoothing: The
  state space approach}}.
\newblock \bibinfo{publisher}{Springer}.
\newblock


\bibitem[\protect\citeauthoryear{Kuznetsov and Mariet}{Kuznetsov and
  Mariet}{2019}]%
        {Kuznetsov2019}
\bibfield{author}{\bibinfo{person}{Vitaly Kuznetsov} {and}
  \bibinfo{person}{Zelda Mariet}.} \bibinfo{year}{2019}\natexlab{}.
\newblock \showarticletitle{Foundations of Sequence-to-Sequence Modeling for
  Time Series}.
\newblock \bibinfo{journal}{\emph{AISTATS}} (\bibinfo{year}{2019}).
\newblock


\bibitem[\protect\citeauthoryear{Larson, Simchi-Levi, Kaminsky, and
  Simchi-Levi}{Larson et~al\mbox{.}}{2001}]%
        {larson01}
\bibfield{author}{\bibinfo{person}{Paul~D. Larson}, \bibinfo{person}{David
  Simchi-Levi}, \bibinfo{person}{Philip Kaminsky}, {and} \bibinfo{person}{Edith
  Simchi-Levi}.} \bibinfo{year}{2001}\natexlab{}.
\newblock \showarticletitle{Designing and manging the supply chain}.
\newblock \bibinfo{journal}{\emph{Journal of Business Logistics}}
  \bibinfo{volume}{22}, \bibinfo{number}{1} (\bibinfo{year}{2001}),
  \bibinfo{pages}{259--261}.
\newblock


\bibitem[\protect\citeauthoryear{Li, Prakash, and Faloutsos}{Li
  et~al\mbox{.}}{2010}]%
        {Li:2010}
\bibfield{author}{\bibinfo{person}{Lei Li}, \bibinfo{person}{B.~Aditya
  Prakash}, {and} \bibinfo{person}{Christos Faloutsos}.}
  \bibinfo{year}{2010}\natexlab{}.
\newblock \showarticletitle{Parsimonious Linear Fingerprinting for Time
  Series}.
\newblock \bibinfo{journal}{\emph{Proc. VLDB Endow.}} \bibinfo{volume}{3},
  \bibinfo{number}{1-2} (\bibinfo{date}{Sept.} \bibinfo{year}{2010}).
\newblock


\bibitem[\protect\citeauthoryear{Mishra, Rohaninejad, Chen, and Abbeel}{Mishra
  et~al\mbox{.}}{2018}]%
        {mishra2018a}
\bibfield{author}{\bibinfo{person}{Nikhil Mishra}, \bibinfo{person}{Mostafa
  Rohaninejad}, \bibinfo{person}{Xi Chen}, {and} \bibinfo{person}{Pieter
  Abbeel}.} \bibinfo{year}{2018}\natexlab{}.
\newblock \showarticletitle{A Simple Neural Attentive Meta-Learner}. In
  \bibinfo{booktitle}{\emph{International Conference on Learning
  Representations}}.
\newblock


\bibitem[\protect\citeauthoryear{Mukherjee, Shankar, Ghosh, Tathawadekar,
  Kompalli, Sarawagi, and Chaudhury}{Mukherjee et~al\mbox{.}}{2018}]%
        {Mukherjee2018}
\bibfield{author}{\bibinfo{person}{Srayanta Mukherjee},
  \bibinfo{person}{Devashish Shankar}, \bibinfo{person}{Atin Ghosh},
  \bibinfo{person}{Nilam Tathawadekar}, \bibinfo{person}{Pramod Kompalli},
  \bibinfo{person}{Sunita Sarawagi}, {and} \bibinfo{person}{Krishnendu
  Chaudhury}.} \bibinfo{year}{2018}\natexlab{}.
\newblock \showarticletitle{{ARMDN:} Associative and Recurrent Mixture Density
  Networks for eRetail Demand Forecasting}.
\newblock \bibinfo{journal}{\emph{CoRR}}  \bibinfo{volume}{abs/1803.03800}
  (\bibinfo{year}{2018}).
\newblock


\bibitem[\protect\citeauthoryear{Oliva, P{\'{o}}czos, and Schneider}{Oliva
  et~al\mbox{.}}{2017}]%
        {OlivaPS17}
\bibfield{author}{\bibinfo{person}{Junier~B. Oliva},
  \bibinfo{person}{Barnab{\'{a}}s P{\'{o}}czos}, {and} \bibinfo{person}{Jeff~G.
  Schneider}.} \bibinfo{year}{2017}\natexlab{}.
\newblock \showarticletitle{The Statistical Recurrent Unit}. In
  \bibinfo{booktitle}{\emph{ICML}}. \bibinfo{pages}{2671--2680}.
\newblock


\bibitem[\protect\citeauthoryear{Qin, Song, Chen, Cheng, Jiang, and
  Cottrell}{Qin et~al\mbox{.}}{2017}]%
        {QinSCCJC17}
\bibfield{author}{\bibinfo{person}{Yao Qin}, \bibinfo{person}{Dongjin Song},
  \bibinfo{person}{Haifeng Chen}, \bibinfo{person}{Wei Cheng},
  \bibinfo{person}{Guofei Jiang}, {and} \bibinfo{person}{Garrison~W.
  Cottrell}.} \bibinfo{year}{2017}\natexlab{}.
\newblock \showarticletitle{A Dual-Stage Attention-Based Recurrent Neural
  Network for Time Series Prediction}. In \bibinfo{booktitle}{\emph{IJCAI}}.
  \bibinfo{pages}{2627--2633}.
\newblock


\bibitem[\protect\citeauthoryear{Quinlan}{Quinlan}{1992}]%
        {quinlan92}
\bibfield{author}{\bibinfo{person}{J.R. Quinlan}.}
  \bibinfo{year}{1992}\natexlab{}.
\newblock \showarticletitle{Learning with continuous classes}.
\newblock \bibinfo{journal}{\emph{Proceedings of the 5th Australian Joint
  Conference on Artificial Intelligence}} (\bibinfo{year}{1992}),
  \bibinfo{pages}{343--348}.
\newblock


\bibitem[\protect\citeauthoryear{Rae, Dyer, Dayan, and Lillicrap}{Rae
  et~al\mbox{.}}{2018}]%
        {rae2018fast}
\bibfield{author}{\bibinfo{person}{Jack~W Rae}, \bibinfo{person}{Chris Dyer},
  \bibinfo{person}{Peter Dayan}, {and} \bibinfo{person}{Timothy~P Lillicrap}.}
  \bibinfo{year}{2018}\natexlab{}.
\newblock \showarticletitle{Fast Parametric Learning with Activation
  Memorization}.
\newblock \bibinfo{journal}{\emph{arXiv preprint arXiv:1803.10049}}
  (\bibinfo{year}{2018}).
\newblock


\bibitem[\protect\citeauthoryear{Rangapuram, Seeger, Gasthaus, Stella, Wang,
  and Januschowski}{Rangapuram et~al\mbox{.}}{2018}]%
        {Rangapuram2018}
\bibfield{author}{\bibinfo{person}{Syama~Sundar Rangapuram},
  \bibinfo{person}{Matthias~W Seeger}, \bibinfo{person}{Jan Gasthaus},
  \bibinfo{person}{Lorenzo Stella}, \bibinfo{person}{Yuyang Wang}, {and}
  \bibinfo{person}{Tim Januschowski}.} \bibinfo{year}{2018}\natexlab{}.
\newblock \showarticletitle{Deep State Space Models for Time Series
  Forecasting}.
\newblock In \bibinfo{booktitle}{\emph{Advances in Neural Information
  Processing Systems 31}}, \bibfield{editor}{\bibinfo{person}{S.~Bengio},
  \bibinfo{person}{H.~Wallach}, \bibinfo{person}{H.~Larochelle},
  \bibinfo{person}{K.~Grauman}, \bibinfo{person}{N.~Cesa-Bianchi}, {and}
  \bibinfo{person}{R.~Garnett}} (Eds.). \bibinfo{pages}{7796--7805}.
\newblock


\bibitem[\protect\citeauthoryear{Ravi and Larochelle}{Ravi and
  Larochelle}{2017}]%
        {ravi17}
\bibfield{author}{\bibinfo{person}{Sachin Ravi} {and} \bibinfo{person}{Hugo
  Larochelle}.} \bibinfo{year}{2017}\natexlab{}.
\newblock \showarticletitle{Optimization as a model for few shot learning}. In
  \bibinfo{booktitle}{\emph{ICLR}}.
\newblock


\bibitem[\protect\citeauthoryear{Rei}{Rei}{2015}]%
        {Rei15}
\bibfield{author}{\bibinfo{person}{Marek Rei}.}
  \bibinfo{year}{2015}\natexlab{}.
\newblock \showarticletitle{Online Representation Learning in Recurrent Neural
  Language Models}. In \bibinfo{booktitle}{\emph{Proceedings of the 2015
  Conference on Empirical Methods in Natural Language Processing}}.
\newblock
\urldef\tempurl%
\url{http://aclweb.org/anthology/D/D15/D15-1026.pdf}
\showURL{%
\tempurl}


\bibitem[\protect\citeauthoryear{Santoro, Bartunov, Botvinick, Wierstra, and
  Lillicrap}{Santoro et~al\mbox{.}}{2016}]%
        {SantoroBBWL16}
\bibfield{author}{\bibinfo{person}{Adam Santoro}, \bibinfo{person}{Sergey
  Bartunov}, \bibinfo{person}{Matthew Botvinick}, \bibinfo{person}{Daan
  Wierstra}, {and} \bibinfo{person}{Timothy~P. Lillicrap}.}
  \bibinfo{year}{2016}\natexlab{}.
\newblock \showarticletitle{Meta-Learning with Memory-Augmented Neural
  Networks}. In \bibinfo{booktitle}{\emph{ICML}}. \bibinfo{pages}{1842--1850}.
\newblock


\bibitem[\protect\citeauthoryear{Seeger, Salinas, and Flunkert}{Seeger
  et~al\mbox{.}}{2016}]%
        {Seeger2016}
\bibfield{author}{\bibinfo{person}{Matthias Seeger}, \bibinfo{person}{David
  Salinas}, {and} \bibinfo{person}{Valentin Flunkert}.}
  \bibinfo{year}{2016}\natexlab{}.
\newblock \showarticletitle{Bayesian Intermittent Demand Forecasting for Large
  Inventories}. In \bibinfo{booktitle}{\emph{Proceedings of the 30th
  International Conference on Neural Information Processing Systems}}
  \emph{(\bibinfo{series}{NIPS'16})}.
\newblock


\bibitem[\protect\citeauthoryear{Shankar and Sarawagi}{Shankar and
  Sarawagi}{2018}]%
        {Shankar2018}
\bibfield{author}{\bibinfo{person}{Shiv Shankar} {and} \bibinfo{person}{Sunita
  Sarawagi}.} \bibinfo{year}{2018}\natexlab{}.
\newblock \showarticletitle{Labeled Memory Networks for Online Model
  Adaptation}. In \bibinfo{booktitle}{\emph{{AAAI}}}.
\newblock


\bibitem[\protect\citeauthoryear{Wen, Torkkola, and Narayanaswamy}{Wen
  et~al\mbox{.}}{2017}]%
        {wen2017multi}
\bibfield{author}{\bibinfo{person}{Ruofeng Wen}, \bibinfo{person}{Kari
  Torkkola}, {and} \bibinfo{person}{Balakrishnan Narayanaswamy}.}
  \bibinfo{year}{2017}\natexlab{}.
\newblock \showarticletitle{A Multi-Horizon Quantile Recurrent Forecaster}.
\newblock \bibinfo{journal}{\emph{arXiv preprint arXiv:1711.11053}}
  (\bibinfo{year}{2017}).
\newblock


\bibitem[\protect\citeauthoryear{Yu, Rao, and Dhillon}{Yu
  et~al\mbox{.}}{2016}]%
        {yu2016temporal}
\bibfield{author}{\bibinfo{person}{Hsiang-Fu Yu}, \bibinfo{person}{Nikhil Rao},
  {and} \bibinfo{person}{Inderjit~S Dhillon}.} \bibinfo{year}{2016}\natexlab{}.
\newblock \showarticletitle{Temporal regularized matrix factorization for
  high-dimensional time series prediction}. In
  \bibinfo{booktitle}{\emph{Advances in neural information processing
  systems}}. \bibinfo{pages}{847--855}.
\newblock


\bibitem[\protect\citeauthoryear{Zhang, Lin, Song, and Dhillon}{Zhang
  et~al\mbox{.}}{2018}]%
        {ZhangLSD18}
\bibfield{author}{\bibinfo{person}{Jiong Zhang}, \bibinfo{person}{Yibo Lin},
  \bibinfo{person}{Zhao Song}, {and} \bibinfo{person}{Inderjit~S. Dhillon}.}
  \bibinfo{year}{2018}\natexlab{}.
\newblock \showarticletitle{Learning Long Term Dependencies via Fourier
  Recurrent Units}. In \bibinfo{booktitle}{\emph{Proceedings of the 35th
  International Conference on Machine Learning, {ICML} 2018,
  Stockholmsm{\"{a}}ssan, Stockholm, Sweden, July 10-15, 2018}}.
  \bibinfo{pages}{5810--5818}.
\newblock


\end{thebibliography}


\end{document}